\documentclass[final,3p,times,twocolumn]{elsarticle}

\usepackage{amsmath,amssymb,amsfonts}

\usepackage[ruled,vlined]{algorithm2e}
\usepackage{algpseudocode}
\usepackage{graphicx}
\usepackage{subcaption}
\usepackage{booktabs}
\usepackage{tabularx}
\usepackage{makecell}
\usepackage{multirow}
\usepackage{ifthen} 
\usepackage{textcomp}
\usepackage{comment}
\usepackage{threeparttable}
\usepackage{xcolor}
\usepackage{adjustbox}
\usepackage{hyperref}
\usepackage{cleveref}

\journal{Robotics and Autonomous Systems}

\begin{document}

\begin{frontmatter}



\title{Path Planning Optimisation for SParse, AwaRe and
Cooperative Networked Aerial Robot Teams (SpArC-NARTs): Optimisation Tool and Ground Sensing Coverage Use Cases}


\author[inst1,inst3,inst2]{Maria Inês Conceição\corref{cor1}}
\ead{ines.conceicao@tecnico.ulisboa.pt}
\author[inst2]{António Grilo}
\ead{antonio.grilo@inov.pt}
\author[inst3]{Meysam Basiri}
\ead{meysam.basiri@tecnico.ulisboa.pt}

\address[inst1]{INESC ID–Instituto de Engenharia de Sistemas e Computadores: Investigação e Desenvolvimento, Instituto Superior Técnico, Universidade de Lisboa, Lisbon, 1000-029, Portugal}
\address[inst2]{INESC INOV, Instituto Superior Técnico, Universidade de Lisboa, Lisbon, 1000-029, Portugal}
\address[inst3]{Institute for Systems and Robotics, Instituto Superior Técnico, Universidade de Lisboa, Lisbon, 1049-001, Portugal}
\cortext[cor1]{Corresponding author}
\begin{abstract}
A networked aerial robot team (NART) comprises a group of agents (e.g., unmanned aerial vehicles (UAVs), ground control stations, etc.) interconnected by wireless links. Inter-agent connectivity, even if intermittent (i.e., sparse), enables data exchanges between agents and supports cooperative behaviours in several NART missions. It can benefit online decentralised decision-making and group resilience, particularly when prior knowledge is inaccurate or incomplete. These requirements can be accounted for in the offline mission planning stages to incentivise cooperative behaviours and improve mission efficiency during the NART deployment. This paper proposes a novel path planning tool for Sparse, Aware, and Cooperative NARTs (SpArC-NARTs) in exploration missions. It simultaneously considers different levels of prior information regarding the environment, limited agent energy, sensing, and communication, as well as distinct NART constitutions. The communication model takes into account the limitations of user-defined radio technology and physical phenomena. The proposed tool aims to maximise the mission goals (e.g., finding one or multiple targets, covering the full area of the environment, etc.), while cooperating with other agents to reduce agent reporting times, increase their global situational awareness (e.g., their knowledge of the environment, including redundant storage for reliability purposes), and facilitate mission replanning, if required. The developed cooperation mechanism leverages soft-motion constraints and dynamic reward shaping based on the Value of Movement and expected communication availability between the agents at each time step. The capabilities of this tool were illustrated with a ground sensing coverage use case. The performance of the proposed mechanism was analysed for different NART constitutions and task distributions. Their performance was compared to one non-cooperative baseline and two cooperative baselines. An ablation study to examine the impact of reward components and connectivity requirements on NART performance and an analysis of the impact of weighting UAV tasks (exploration and reporting) differently were also performed.

\end{abstract}




\end{frontmatter}


\section{Introduction}
\begin{figure*}[!t]
    \centering
    \includegraphics[width=\linewidth]{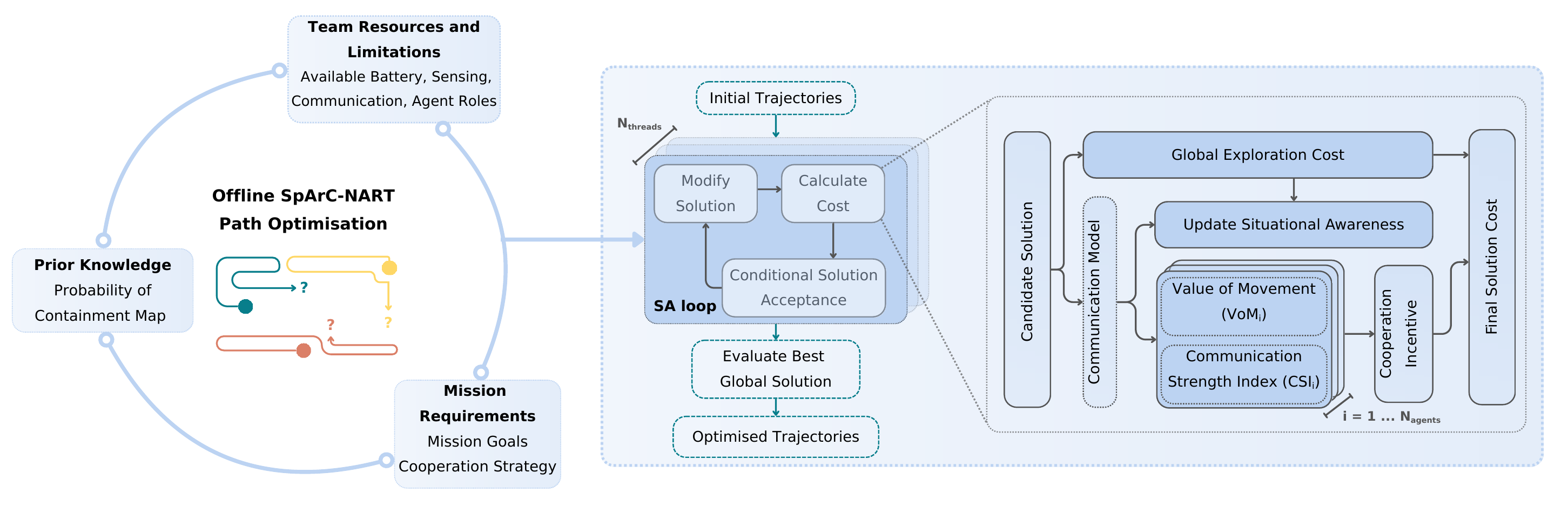}
    \caption{Offline SpArC-NART Path Planning Optimisation Design Framework. Agent trajectories are optimised with Simulated Annealing (SA). The length of each SA loop will depend on user-defined parameters. Used parameters and conditional acceptance will be provided in Section \ref{CPPUseCase}. The cost of each candidate solution considers the exploration and communication performance. }
    \label{overall}
\end{figure*}

In networked aerial robot teams (NARTs), a team of Unmanned Aerial Vehicles (UAVs) is connected by wireless links that support data exchange \cite{CMAMCA}. Optionally, the UAVs may also be interlinked to external entities, such as humans and ground control stations (GCSs). Existing links facilitate data sharing between agents, providing crucial updates on mission status and enhancing their local environmental knowledge. Cooperative NART agents can leverage their diverse characteristics, including varying computational and onboard sensing capabilities, as well as agent mobility, to enhance the overall capabilities of the group. This cooperative behaviour helps establish a shared situation awareness between the agents, which supports decentralised decision-making and mission efficiency \cite{CHANDRAN2024100858}, \cite{9920737}.

Ground communication \cite{9825683}, \cite{10579859} and sensing coverage \cite{apostolidis2022cooperative}, target search and tracking \cite{10433718}, \cite{10021886}, and environment exploration and mapping \cite{9297053},  are among the most frequent applications of NARTs in recent literature. The efficiency of these applications benefits significantly from the implementation of cooperation strategies. However, the dependence on stable communication introduces a set of additional difficulties to their successful implementation. When the optimisation of a mission requires the dispersion of agents in the environment, for example, to explore a vast area which extends far beyond the individual abilities of the agents, direct communication between agents may not be possible. Alternative approaches, which focus on maintaining inter-agent connectivity \cite{10039398}, can also considerably restrict the movement of the NART when it comprises a small number of agents. To increase the flexibility of the group, additional agents can also be introduced into the NART, such as data mules \cite{datamule}, \cite{s24196218} or relay agents in a chain formation\cite{141}. If adding an indefinite number of agents is not feasible, alternative approaches that target a balance between communication and mission development must be considered. 

Overall, data exchanges enable the relay of mission and environment updates through the NART agents. Agent reporting also facilitates knowledge sharing during mission development, thereby reducing uncertainty about the environment and increasing the situational awareness of the agents. It also enhances NART resilience. The cooperation opportunities can be leveraged for online mission replanning, if necessary. These can also help to overcome limitations arising from incomplete or inaccurate prior knowledge and to prioritise agent reporting.

Agent-limited battery and sensing restrictions further aggravate the contradiction between the physical proximity required for agent reporting and the physical distancing needed to meet the exploration requirements. Thus, when mission goals require both, it is important to establish an initial plan which targets exploration efficiency while still providing a high probability of cooperation. The proposed offline optimisation aims to address this issue. 

The proposed work focuses on offline trajectory optimisation for exploration missions, particularly informative ground sensing coverage and target search. It simultaneously considers team resources and limitations, prior knowledge and mission requirements, as illustrated in Fig. \ref{overall}. The path of each NART agent is optimised based on its limited energy, sensing uncertainty, different levels of prior knowledge, and estimated communication with other agents. The developed work proposes a flexible approach to cooperation, which leverages a dynamic trade-off between exploration and reporting under intermittent communication to maximise exploration efficiency while providing opportunities for cooperation.

Instead of choosing between ensuring a minimal communication Quality of Service (QoS) between the agents \cite{8691430} or maximising an exploration mission goal \cite{OriginalPP}, the proposed dynamic cooperative incentives aim to  balance between them during the path optimisation. Furthermore, instead of solely relying on undefined communication ranges \cite{121},\cite{11083772} or euclidean distance-based criteria for communication \cite{124}, \cite{10679913}, the proposed approach considers the limitations of a user-selected radio technology to estimate inter-agent connectivity. 

Estimated connectivity and dynamic reporting urgency are encompassed as reward shapping components of cooperation. These softly defined rewards replace hard constrained reporting intervals, explicit scheduling of meeting events, meeting points and participant optimisation. With the proposed approach, meeting events involving specific agents are more predictable when the cooperation mechanism  (e.g., planned cooperation) is taken into account in the path optimisation. Two main NART constitutions were considered: homogeneous (consisting only of UAVs) and heterogeneous (UAVs supported by external entities, whether static or mobile).

The contributions of this paper can be summarised as follows:

\begin{itemize}
  \item A problem formulation for sparse, communication-limited NART missions, under energy, sensing and prior knowledge limitations.
  \item A communication awareness module for mission planning, considering restrictions of user-specified radio technology and physical phenomena (e.g., signal fading) with easy integration in NART development and a radio-aware connectivity index ($CSI$) output as a higher fidelity metric for inter-agent connectivity.
  \item A reward-based planned cooperation mechanism which encourages rendezvous-style data exchanges while still promoting exploration, using radio-aware connectivity index and reporting urgency as reward shaping components. 
  \item An offline trajectory optimisation, ablation study and evaluation across NART team compositions, cooperation strategies and mission environments.
\end{itemize}

This paper addresses the Related Work in Section \ref{related_work}. The proposed methods are presented in Section \ref{SpArC-NARTs}. The definition of the Ground Sensing Coverage use case is presented in Section \ref{CPPUseCase}. Section \ref{sim_Results} presents the simulation results. Section \ref{CMCA_App} discusses the communication and motion coordination awareness of the proposed approach. Section \ref{conclusion} concludes the paper.

\section{Related Work}\label{related_work}

Joint integration of exploration and reporting in NART missions has been considered for large teams \cite{120} and for NARTs with permanently connected agents \cite{10440144}, even with relaxed data-exchange requirements \cite{10801613}, \cite{dppm}. 

\cite{10161401} addresses the challenge of minimising the motion restriction resulting from maintaining line-of-sight (LOS) communication between cooperative agents in an exploration mission. A centralised approach was proposed to maintain a minimal set of connected agents. A large team of robots with parallel tasks was considered. The number of links in each step to maintain the group interconnected is determined through a communication-constrained minimum spanning tree. The NART topology adjustments ensure both global and local LOS connectivity, increasing NART flexibility and minimising the impact of communication on the robot tasks.

\cite{10802832} adopts a different perspective, focusing on decentralised communication-dependent cooperation within a leader-follower group strategy. The leader robot follows an exploration policy, and the followers have a policy network. The latter leverages a centralised training and decentralised execution (CTDE) strategy that combines imitation learning (IL) and reinforcement learning (RL). The first aims to maintain connectivity to the follower, and the second is used for improved decision-making. A decentralised POMDP-based formulation rewards new areas discovered by the followers while penalising the loss of connectivity between them. Permanent direct or multi-hop pairwise communication is aimed at. The communication range was considered an euclidean distance between the agents.

Droppable radios \cite{9837416} have also been introduced to a multi-robot exploration mission to support further inter-robot communication and therefore cooperation as external passive agents. The impact of limited communication in multi-robot exploration has also been addressed through the development of modules to reduce the bandwidth requirements for communication \cite{10876135}. The developed work was also able to reduce the redundancy in exchanged map data.

Ensuring permanent connectivity in the deployment of a small-sized NART in an environment which extends far beyond the sensing ability of the agents is a challenging task. Considering intermittent communication between NART agents can reduce the motion restriction arising from permanent connectivity and improve mission performance. It can be particularly advantageous to deploy a small team of agents in large-scale exploration missions. The cooperation resulting from the intermittent inter-agent links enables sensor data sharing and opportunities to report mission-relevant updates, consequently improving global situational awareness, decision-making, and adaptability to unforeseen mission and environmental updates, while still promoting the agent exploration. These exchanges can occur in an opportunistic or planned approach. 

Opportunistic cooperation occurs when the communication requirements between two agents are met, thereby enabling data exchange, as in \cite{144}. As no specific incentive for cooperation is given, these exchanges are not guaranteed to occur, unless the size of the NART team is defined proportionally to the dimension of the environment that needs to be explored \cite{26}. In this case, opportunistic data exchanges are guaranteed to occur, even in uncertain times and locations, as in \cite{119}. 

Planned cooperation, on the other hand, involves methods to enhance the predictability of meeting events during mission development. Offline strategies can account for initial knowledge of the environment and agent limitations to provide an initial optimisation of the mission. Online strategies can further enhance cooperation efficiency by reacting to perceived inter-agent communication.

In the proposed work, opportunistic and planned cooperation is considered. A dynamic reward-based approach is used to induce proximity or distancing between NART agents. The spatial proximity indirectly creates opportunities (e.g., meeting points or rendezvous) for direct and indirect reporting between the agents of the NART. The distancing of the agents promotes the exploration of the environment. The rewards are calculated at each time step for each agent. The proposed approach simultaneously considers the time elapsed since the previous data exchange with an exponentially growing weight and communication requirements. The limitations of the user-selected radio technology define the later requirements.

\subsubsection*{Multi-Robot Planned Cooperation}
 
Inter-agent cooperation can be planned by explicitly defining one or multiple meeting points. The meeting points are transmitted to the agents, including the location, time, and specific agents who will participate in the exchange. A planned cooperation approach can also be defined indirectly. While optimising other mission goals, the proximity between agents can allow data sharing, even without an explicit meeting point definition.

Considering a single meeting point (explicitly defined) can require additional strategies to manage reporting between agents and define subsequent tasks for each agent \cite{141}. This procedure must account for the scenario in which several agents travel to the meeting point simultaneously, as well as when an agent is waiting at the meeting point. Optimising the definition of explicit meeting points can also help minimise the risk of jeopardising mission performance. Particularly, if travelling to the meeting point restricts the ability of the agents to explore other areas of the environment and contribute further to the mission goal (e.g., finding multiple targets or mapping the environment).

Rendezvous-based approaches can improve exploration efficiency in unknown environments \cite{10679913}, compared to m-TSP approaches such as \cite{10106111}. An online rendezvous is regarded in this work as an event that gathers all agents at a specific location in the environment. The rendezvous point is explicitly defined as the point in the environment to which the agents can travel the fastest. During a rendezvous, the agents cooperatively combine local lightweight feature-based hybrid topological maps (FHT\_Map) to achieve faster exploration. When not in a rendezvous, robots perform a Next Best View (NBV)-based exploration, estimate relative positions (RPs) of other agents and update the Voronoi-based space partitions of the environment. The rendezvous point is only calculated once the RPs of all agents are determined. Limited communication impacts the  exchanges that support the meeting point definition. Connectivity requirements include a limited bandwidth. However, it is considered in an unlimited range.

An online explicit rendezvous definition was also considered in \cite{9981898} for a multi-robot exploration mission. This work defined a state machine with four main behaviours (i.e., explore, rendezvous, search, and exploit) and time or event-based interconnections. All agents start the mission by exploring the environment in a fully connected network. A rendezvous is considered to occur when all agents are connected. It was highlighted to improve situational awareness and allocate tasks among the agents. The meeting point is also dynamically computed to minimise the travelling distance for all participants. In case a robot fails to rendezvous with other agents, it will return to exploration. The communication range that allows data sharing between agents during a rendezvous is, however, considered an euclidean distance. The effect of obstacles and the capabilities of a specific radio technology is not taken into account.

\begin{table*}[!t]
\centering
\caption{Comparison of Related Work on Multi-Robot Cooperation.}
\label{tab:comparison}
\resizebox{\textwidth}{!}{%
\renewcommand{\arraystretch}{1.5}
\begin{tabular}{|c|c|c|c|c|c|c|c|c|c|c|c|}
\hline
\multirow{2}{*}{\raisebox{-0.5\normalbaselineskip}{\textbf{Reference}}} & \multirow{2}{*}{\raisebox{-0.5\normalbaselineskip}{\textbf{Approach}}} & \multirow{2}{*}{\raisebox{-0.5\normalbaselineskip}{\parbox{2cm}{\centering\textbf{Permanent Connectivity}}}} & \multicolumn{3}{c|}{\textbf{Cooperation}} & \multirow{2}{*}{\parbox{2.5cm}{\centering\vspace{0.3cm}\textbf{Team Composition}\vspace{0.3cm}}} & \multirow{2}{*}{\parbox{1.5cm}{\centering\vspace{0.15cm}\textbf{Integrated Prior Knowledge}\vspace{0.25cm}}} & \multicolumn{4}{c|}{\textbf{Link Connectivity Range}} \\
\cline{4-6} \cline{9-12}
 &  &  & \parbox{2cm}{\centering Multiple Meeting Point} & Definition & Participants & &  & Limited & \parbox{1.3cm}{\centering Euclidean Distance} & \parbox{1.5cm}{\centering Radio Technology} & \parbox{1.5cm}{\centering Physical Phenomena} \\
\hline
{\cite{141}} & Online & No & No & Explicit & Pairwise & Homogeneous &  Yes & Yes & Yes & No & No \\
\hline
{\cite{10679913}} & Online & No & Yes & Explicit & Global & Heterogeneous & No & Yes & Yes & No & No \\
\hline
{\cite{9981898}} & Online & No & Yes & Explicit & Global & Homogeneous & No & Yes & Yes & No & No \\
\hline
{\cite{10802832}} & Online & Yes & N/A & N/A & N/A & Homogeneous &No & Yes & Yes & No & No \\
\hline
\textbf{Proposed Work} & \textbf{Offline} & \textbf{No} & \textbf{Yes} & \textbf{Indirect} & \textbf{Pairwise} & \parbox{2.5cm}{\centering \textbf{Homogeneous and Heterogeneous}} & \textbf{Yes} & \textbf{Yes} & \textbf{No} & \textbf{Yes} & \textbf{Yes} \\
\hline
\end{tabular}%
}
\end{table*}

Indirectly promoting pairwise rendezvous between agents has been addressed by assigning specific tasks, such as explorers, data mules, or relays. The path of the agents can be optimised prior to mission start accordingly. The path of the explorers focuses on environment coverage performance. The path of the relays is optimised to improve reporting between explorers \cite{10117548}. The role of data mules is also highlighted to extend terrestrial network coverage and data collection \cite{moheddine2023uav}. 

The work \cite{LUPERTO2025105137} promotes rendezvous-based cooperation in an unknown indoor environment multi-robot exploration mission. A rendezvous is considered an online task that requires physical proximity, similar to a flight in formation. Communication requirements are also considered with distance-based connectivity under LOS conditions, direct and multi-hop links between agents. While agents start the mission with asynchronous exploration, clusters of connected agents can establish leader-follower dynamics. In this work, the rendezvous cost is associated with the time required for agents to perform the task. Facilitating the rendezvous can reduce the associated cost, the length of the trajectories and the overlap between them. Agent backtracking to previously explored areas and motion incentives for high-connectivity areas, where rendezvous are more likely to occur (e.g., corridors and hallways), were considered. The performance of reporting or situational awareness was not addressed. Ultimately, the area explored by the group of robots was successfully maintained using the proposed strategy. Some settings noted, however, a decrease in this metric.

Overall, initial mission plans consider assumptions about inter-agent communication, sensing uncertainty, and available agent energy. The NART deployment often considers a high probability of opportunistic data exchanges. The chosen mission environments do support these assumptions, as the number of agents is often adequate for the environment dimensions. Such is not assumed in the proposed work. Additional hard constraints for reporting are also often implemented alongside inter-agent distance as connectivity requirements. In the proposed work, the offline optimisation aims to maximise mission goals while also accounting for communication opportunities under the expected limitations of the implemented radio technology and signal fading. Impact on global situational awareness is monitored as well.

The combination of exploration and inter-agent communication has been addressed in the literature, particularly in online settings, as presented in Table \ref{tab:comparison}. The current work follows a different approach, focusing on offline optimisation to provide an initial mission strategy tuned to a set of requirements (e.g., exploration and reporting). These can be further improved in a second stage based on online-perceived communication conditions and mission development. For example, online reporting between agents can be used as an opportunity for path replanning. This requirement for frequent rendezvous can be accommodated through offline procedures. For this reason, it is important to analyse reporting, global situational awareness, and exploration performance, as in the proposed work. The resulting strategy can thus increase the resilience of the NART to unforeseen emergencies, dynamic environments and incorrect prior knowledge. This approach also helps to overcome a frequent limitation of offline multi-robot mission planning \cite{10102336}. To the best of the author's knowledge, the proposed work provides a more robust cooperation strategy for exploration missions in large environments by small NART teams (e.g., homogeneous and heterogeneous) under intermittent connectivity, leveraging implicitly defined cooperation and radio technology-aware connectivity constraints.

\subsubsection*{Ground Sensing Coverage}
A ground sensing coverage mission was chosen as a use case in this paper. Informative path planning and coverage path planning strategies are frequently used in this setting. The mission goals often involve exploring an environment while covering the largest possible area. Prior environment knowledge can also be used to maximise exploration performance and increase the probability of finding targets in the environment. 

The offline information-aware coverage path planning work proposed by \cite{OriginalPP} does not consider inter-agent cooperation. It focuses on maximising the ability of agents to explore the space and find a target in the shortest amount of time under energy and sensing limitations. This work also compares different path optimisation strategies for an efficient mission, balancing expected performance with optimisation convergence time.

\cite{10534782} deploys a heterogeneous team with varying sensing abilities for an online contamination mapping mission (e.g., pollution in a lake). The trajectories of the agents are optimised through informative-path planning to build a contamination map with minimal errors. The acquisition of key information regarding contamination is maximised, while the errors of the model concerning the real information are minimised. Agent decision-making leverages a single centralised double deep Q-learning strategy to increase estimated agent future rewards based on observed ones, under a collision-free trajectory. While agents cooperate towards a common goal (i.e., contamination map), the communication-based cooperation and its limitations are not addressed.

A multi-robot group implements a leader-follower strategy to create a map of physical values of interest in an environment in \cite{9561955}. Each follower is assigned to a disjoint area of interest, defined by the leader. Independently, each follower maximises the information gain in a collision-free trajectory. Delay-tolerant networking is considered to enable data sharing and map definition within a communication range. Pairwise data exchanges are expected to occur in this setting. However, this work does not address radio technology and energy limitations and their impact on mission performance. This online strategy considers mission replanning to minimise deviations between sampled data and predicted results. The leader can compute new areas to visit if the information quality does not meet a defined quality threshold.

\begin{table*}[!t]
\centering
\caption{Comparison of Related Work for Ground Sensing Coverage.}
\label{tab:extended_comparison}
\resizebox{\textwidth}{!}{%
\renewcommand{\arraystretch}{1.5}
\begin{tabular}{|c|c|c|c|c|c|c|c|c|c|c|c|c|c|c|c|}
\hline
\multirow{3}{*}{\raisebox{-1.0\normalbaselineskip}{\textbf{Reference}}} & \multirow{3}{*}{\raisebox{-1.0\normalbaselineskip}{\parbox{1.5cm}{\centering\textbf{Approach}}}} & \multirow{3}{*}{\raisebox{-1.0\normalbaselineskip}{\textbf{Cooperation}}} & \multirow{3}{*}{\raisebox{-1.0\normalbaselineskip}{\parbox{2.5cm}{\centering\textbf{Team\\Composition}}}} & \multirow{3}{*}{\raisebox{-1.0\normalbaselineskip}{\parbox{2cm}{\centering\textbf{Task\\Assignment}}}} & \multirow{3}{*}{\raisebox{-1.0\normalbaselineskip}{\parbox{1.8cm}{\centering\textbf{Global\\Connectivity}}}} & \multicolumn{7}{c|}{\textbf{Agent Limitations}} & \multicolumn{3}{c|}{\multirow{2}{*}{\textbf{Performance Analysis}}} \\
\cline{7-13}
 & & & & & & \multicolumn{4}{c|}{\textbf{Link Connectivity Range}} & \multirow{2}{*}{\parbox{1.5cm}{\centering\textbf{Integrated \\ Energy}}} & \multirow{2}{*}{\parbox{1.5cm}{\centering\textbf{Integrated \\ Sensing}}} & \multirow{2}{*}{\parbox{1.5cm}{\centering\textbf{Integrated \\ Prior\\Knowledge}}} & \multicolumn{3}{c|}{} \\
\cline{7-10} \cline{14-16}
 & & & & & & \parbox{1.8cm}{\centering\textbf{Limited}} & \parbox{1.5cm}{\centering\textbf{Euclidean\\Distance}} & \parbox{1.5cm}{\centering\textbf{Radio\\Technology}} & \parbox{1.5cm}{\centering\textbf{Physical\\Phenomena}} & & & & \textbf{Reporting} & \parbox{1.8cm}{\centering\textbf{Global\\Situational\\Awareness}} & \parbox{1.8cm}{\centering\textbf{Optimised\\Path and\\Main Goal}} \\
\hline
\cite{OriginalPP} & Offline & No & Homogeneous & Explicit & N/A & N/A & N/A & N/A & N/A & Yes & Yes & Yes & No & No & Yes \\
\hline
\cite{10534782} & Online & Yes & Heterogeneous & Explicit & N/A & N/A & N/A & N/A & N/A & Yes & Yes & Yes & No & Yes & Yes \\
\hline
\cite{9561955} & Online & Yes & Homogeneous & Explicit & Yes & Yes & Yes & No & No & No & Yes & Yes & No & Yes & Yes \\
\hline
\cite{10610484} & Offline & No & Homogeneous & Explicit & N/A & N/A & N/A & N/A & N/A & No & Yes & Yes & No & No & Yes \\
\hline
\cite{9488669} & Online & Yes & Homogeneous & Explicit & N/A & N/A & N/A & N/A & N/A & Yes & No & Yes & No & No & Yes \\
\hline
\cite{10742370} & Offline & Yes & Heterogeneous & Explicit & Yes & Yes & Yes & No & No & No & Yes & Yes & No & No & Yes \\
\hline
\cite{10420441} & Offline & Yes & Homogeneous & Explicit & Yes & Yes & Yes & No & No & No & Yes & Yes & No & No & Yes \\
\hline
\cite{8794090} & Online & Yes & Homogeneous & Explicit & Yes & Yes & Yes & No & No & No & Yes & Yes & No & No & Yes \\
\hline
\cite{10297577} & Online & Yes & Homogeneous & Explicit & Yes & Yes & Yes & No & No & Yes & No & No & No & No & Yes \\
\hline
\cite{10852549} & Online & Yes & Heterogeneous & Explicit & N/A & N/A & N/A & N/A & N/A & Yes & No & No & No & No & Yes \\
\hline
\cite{11083772} & Online & Yes & Homogeneous & Explicit & N/A & N/A & N/A & N/A & N/A & Yes & No & No & No & No & Yes \\
\hline
\textbf{Our Work} & \textbf{Offline} & \textbf{Yes} & \parbox{2.5cm}{\centering\textbf{Homogeneous and\\Heterogeneous}} & \parbox{2cm}{\centering\textbf{Explicit and/or\\Implicit}} & \textbf{No} & \textbf{Yes} & \textbf{No} & \textbf{Yes} & \textbf{Yes} & \textbf{Yes} & \textbf{Yes} & \textbf{Yes} & \textbf{Yes} & \textbf{Yes} & \textbf{Yes} \\
\hline
\end{tabular}%
}
\end{table*}

\cite{10610484} also addresses multi-robot informative path planning (MIPP). The problem is framed as an offline sensing placement optimisation and a following visiting order optimisation for environment monitoring. The first challenge aims to maximise mutual information, while the second addresses the impact of routing constraints (i.e.,distance budget and velocity limits). The impact of cooperation was not considered.

The performance of deep RL-based cooperation was also tested in an MIPP formulation for indoor environmental monitoring \cite{9488669}. A WiFi Received Signal Strength collection was used as a use case. Independent learning through credit assignment and sequential rollout-based learning were two cooperative strategies compared. In the first approach, cooperation considers joint action states among agents and team-shared rewards. The second approach considers sequential robot planning. The update of the current Q-function of an agent considers the subsequent actions of the other agents at step $s$. RL-based approaches were overall more efficient than the genetic algorithm-based Baseline approaches. Communication requirements for cooperation and mission development were not addressed in this work.

Other variations of the MIPP problem have also jointly addressed information gain with dynamic topology control for heterogeneous robots \cite{10742370} and permanent connectivity maintenance \cite{10420441}, \cite{8794090}.  \cite{10420441} considers connectivity between a homogeneous robot team with 10 agents. \cite{8794090} considers constrained bipartite graph matching with minimal node separators and robot path allocation. Overall, communication requirements take into account inter-agent distance criteria. Sensor uncertainty and energy limitations are also often disregarded.  The implementation of reinforcement learning approaches has also highlighted the benefits of global-and-local attention mechanisms to achieve multi-UAV coordination under a fully distributed action setting \cite{10297577}, extrinsic-and-intrinsic rewards to boost cooperation upon target detection and encirclement \cite{10852549}, as well as dual-centralised Q-networks to improve UAV trajectories safety in a cooperative path planning mission\cite{11083772}. These approaches could support MIPP missions. Nevertheless, some of the previous assumptions are kept. \cite{10297577} considered euclidean distance as an indication for connectivity and abstracted energy and sensing limitations of the UAVs. \cite{10852549} and \cite{11083772} considered limited UAV energy, environment characteristics and responses to online events (e.g., avoiding collisions with dynamic obstacles and the impact of climatic conditions) and abstracted communication limitations.

Table \ref{tab:extended_comparison} resumes the analysed related work on ground sensing coverage. It highlights the considerations of several recent works on NART agents, cooperation strategies, agent limitations, and performance analysis. This table also highlights the novelty of the proposed work. To the best of the author's knowledge, the proposed work is novel, as it includes softly defined dynamic cooperation rewards within a multi-robot informative cooperative path planning approach, with single-objective exploration optimisation and communication-aware reward shaping. While agent roles (e.g., exploration and reporting) can be explicitly assigned, agents can also seamlessly interchange between them as a result of the optimisation (i.e., implicit definition). Inter-agent connectivity considers the limitations of radio technology and physical phenomena, rather than solely the euclidean distance between agents. Path optimisation simultaneously considers mission goals, intermittent communication under realistic constraints, energy limitations, prior environment knowledge, sensing uncertainty, and different NART compositions (i.e., homogeneous and heterogeneous), while other works target at most some of these factors. A combination of opportunistic and implicit planned cooperation is considered to optimise mission performance, agent reports, and group resilience to unforeseen occurrences during deployment. For each set of optimised paths, the trade-off between NART reporting, global situational awareness, and coverage path planning performance is also evaluated.

\section{SParse, AwaRe and Cooperative Networked Aerial Robot Teams (SpArC-NARTs)}\label{SpArC-NARTs}

This section introduces SpArC-NARTs, their mission, agent abilities and restrictions. The following characteristics were considered:
\begin{itemize}
    \item Sparse: NART agents are sparsely connected. Such entails that any data exchanges between the agents are not only restricted to the communication limitations but also do not occur at all timesteps. Mission adjustments and situational awareness may not be updated instantly for all agents, but rather after a data exchange has occurred. 
    \item Aware: NART mission is developed under the following considerations:
    \begin{itemize}
        \item Prior Knowledge - The initial knowledge of the environment is represented through a set of probabilities. The probability of one of the targets being in a particular cell will be given by a Probability of Containment (PoC), as presented in \cref{priork}.
        \item Limited UAV Energy - The UAVs that are integrated in the NART have a limited flight time similarly to \cite{OriginalPP}.
        \item Sensor Uncertainty - Onboard sensors have an associated measurement uncertainty. The probability of an agent detecting a target in a cell which indeed contains a target is given by the probability of detection, $pod$. Consequently, the $pod$ represents the accuracy of the sensor.
        \item Limited Communication - Communication between agents is limited by the chosen technology, hardware restrictions and physical effects. 
    \end{itemize}
    \item Cooperative: The NART Agents can cooperate according to different strategies. The roles of relaying information and exploring can be previously assigned to the NART Agents.
\end{itemize}

\subsection{Mission Environment and Prior Knowledge} \label{priork}

The environment is discretised into a grid aligned with the onboard sensor field of view as in \cite{OriginalPP}. The dimension of a cell in the grid is defined so that the onboard sensor can completely cover its area. Each cell has an associated Probability of Containment (PoC) value. This prior may be uniform or non-uniform (i.e., clustered) depending on the application. The grid has an associated inertial frame ${I}$.

\subsection{NART Agents}

Two types of agents were considered: Unmanned Aerial Vehicles (UAVs) and External Entities (EEs). The EE can be either static (S-EE) or mobile (M-EE).  $\mathcal{NART}$ is considered the set of NART agents and $|\mathcal{NART}|$ the number of agents in the NART. Each agent may take on the roles of exploration and/or information relay. 


\subsection*{\textbf{UAV Kinematics}}
\begin{figure}[htbp]
    \centerline{\includegraphics[scale=0.23]{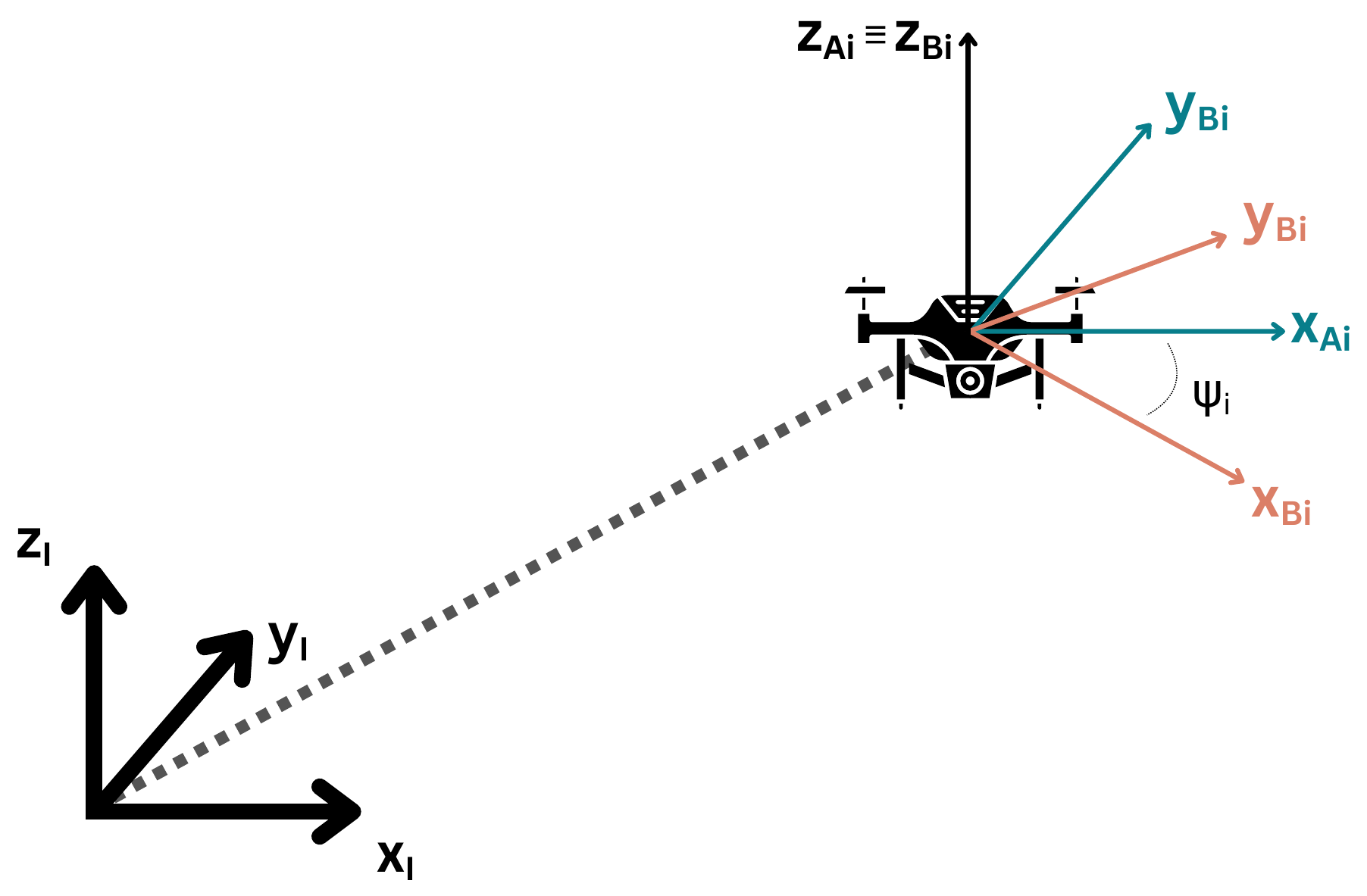}}
    \caption{UAV Kinematics.}
    \label{refs}
\end{figure}

 Each UAV, $UAV_i$, has a reference frame $\{A_i\}$ as a local mobile frame attached to its centre of mass, and a body frame $\{B_i\}$ as a local rotated mobile frame also attached to its centre of mass. $\{A_i\}$ observes a translation motion with respect to $\{I\}$, while $\{B_i\}$ observes translation and rotation motions with respect to $\{I\}$, as illustrated in Fig. \ref{refs}. The rotation is performed according to the rotation matrix $R$, which is presented in Eq. \ref{rot_matrix}. $R_i$ $\in$ SO(3), defines the rotation from $\{B_i\}$ to $\{A_i\}$, where $R_i = I$ if $\psi_i = 0$.

\begin{equation}
    \label{rot_matrix}
{}_{B_i}^{A_i}R_i =
\begin{bmatrix}
\cos \psi_i  & \sin \psi_i  & 0 \\
-\sin \psi_i  & \cos \psi_i  & 0 \\
0 & 0 & 1
\end{bmatrix} .
\end{equation}
The position of each UAV is considered as $\mathbf{p_i} = \begin{bmatrix} x_i & y_i & z_i \end{bmatrix}^{T}$. Moreover, the position $ p_i \in R^{3}$ of the origin of $\{A_i\}$ and $\{B_i\}$ is considered relative to $\{I\}$. As UAVs move along the centre of adjacent cells in the environment, the values $x_i$ and $y_i$ at each step $s$ will take the value of the (x, y) centre coordinate of the cell visited by $UAV_i$. The UAVs were considered to be at distinct altitudes, according to $z_i = \Delta z_{agent}(i+1) [m]$. $\Delta z_{agent}$ is a user-defined parameter that corresponds to the vertical displacement between two UAVs. 
At last, the following parameters were considered:
\begin{itemize}
    \item The linear velocity $v_i$ $\in R^{3}$ of the origin of $\{A_i\}$ and $\{B_i\}$  relative to $\{I\}$;
    \item The yaw angle $\psi_i $ that $\{B_i\}$ observes regarding $\{A_i\}$;
    \item The fixed angular velocity r of $\{B_i\}$ relative to {A}.
\end{itemize}
\subsection*{\textbf{EE Kinematics}}

 The position of the S-EE is defined by the user, resulting in $\mathbf{p_{S-EE}} = \begin{bmatrix} x & y & 0 \end{bmatrix}^{T}$. The M-EE kinematics share overall the considerations presented for the UAV agents. The position of the M-EE, $p_{M-EE}$, is defined according to the mission environment and the number of UAVs. The flight of the M-EE was considered to be $\Delta z_{agent}$[m] above the UAV with the highest altitude. Therefore, it was considered $\mathbf{p_{M-EE}} = \begin{bmatrix} x & y & z \end{bmatrix}^{T}$, where $z = \Delta z_{agent}* |\mathcal{NART}| [m]$ and the parameters $x$ and $y$ are given according to the center coordinates of the cell in which the M-EE is placed at each timestep $s$.

\subsection{Communication Model}

\begin{figure*}[!thpb]
    \centering
    \includegraphics[width=\linewidth]{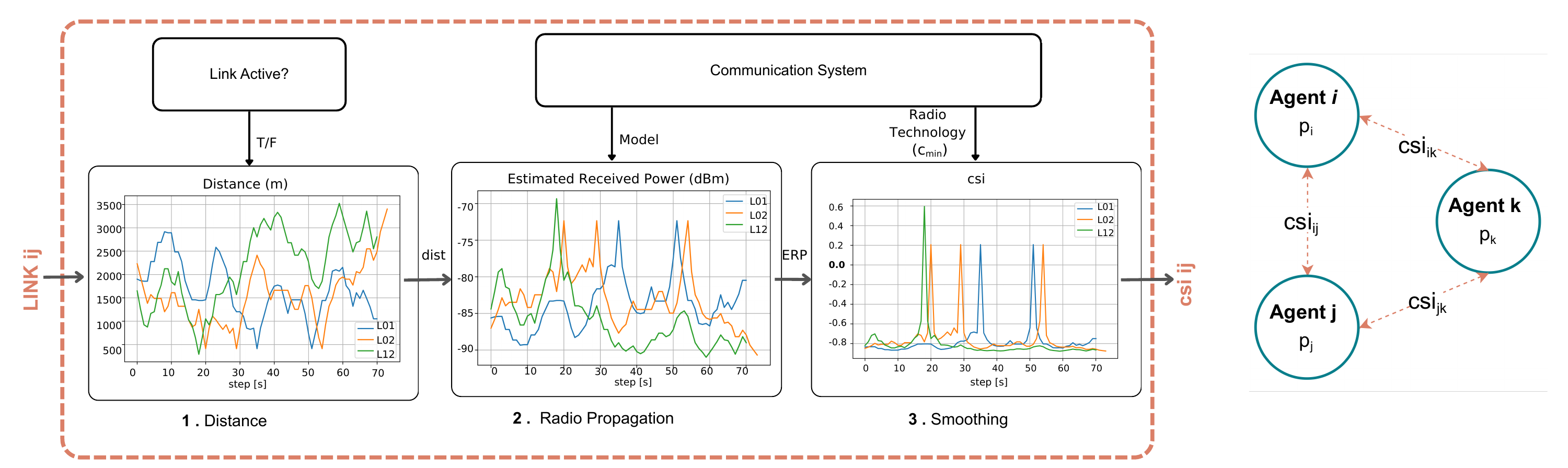}
    \caption{Communication Model - Example for 3 UAVs.}
    \label{CommModuleFig}
\end{figure*}
The proposed communication model estimates the communication link feasibility between two agents given user-selected radio technologies, antenna gains, and a propagation model. The radio technology characteristics influence the expected range of communications. The receiver sensitivity value $c_{min}$ provides the minimum received power required for a receiver to decode a message with an acceptable error packet rate. $c_{min}$ is considered the criteria for a successful a data exchange in this work, instead of the inter-agent distance. Antenna characteristics, such as the radiation pattern, the receiver and transmitter gains, $G_r$ and $G_t$ respectively, the radio technology frequency $f$ and the transmitted power $P_t$ are also taken into account. An isotropic radiation pattern was considered at this stage. 

A communication link (e.g. $l_{01}$) can only be analysed if both agents (e.g. agent 0 and agent 1) are participating in the mission at step $s$. If this condition is met, the link is considered active. The proposed communication model evaluates an active link at a given timestep $s$. Three sequential steps were considered, as illustrated in  Fig. \ref{CommModuleFig}. The first step calculates the inter-agent distance ($dist$, in m) at step $s$. In step 2, the radio signal propagation between two agents (i.e. the transmitter and the receiver) is estimated with a wireless channel model. The Friis Propagation Model was considered in this work. Its output provides the estimated received power ($ERP$) in dBm. To facilitate the usage and integration of the communication model, a smoothing function was added in step 3. The smoothing function is presented in Eq. \ref{eq_smoothing}.  Its output is the communication strength index ($csi\, \in\, [-1;1]$). Link connectivity is thus associated with links with a $csi_{ij} \geq c_{min}$.

\begin{equation} 
    csi_{ij}(s) = \left(
    \frac{
        k \left( ERP_{ij}(s) - (c_{\min} - \varepsilon) \right)
    }{
        1 + \left| k \left( ERP_{ij}(s) - (c_{\min} - \varepsilon) \right) \right|
    }
    \right).
    \label{eq_smoothing}
\end{equation}

The parameter $k$ regulates the curvature of the sigmoid. $\epsilon$ creates a minor displacement of the centre of the function, so that an estimated received power $csi\_{ij}$ equal to $c_{min}$ reflects a successful data exchange. The complete communication model is presented in Alg. \ref{cm}.

\begin{algorithm}[!h]   
\caption{Communication Model - Single Link Example.}
\label{cm}
\SetAlgoLined
\KwIn{$agent_i$, $agent_j$, $p_i$, $p_j$, $c_{min}$, $s$}
\KwOut{$csi$}
\tcp{\scriptsize link be established if agents $i$ and $j$ are in the mission at step $s$}
$link \gets$  \textbf{link\_active(}\text{$agent_i$, $agent_j$, $s$}\textbf{)}\;
\If{$link$}{    
    \tcp{\scriptsize compute distance between receiver and transmitter}
    Compute $dist \gets$ \textbf{euclidean3Ddist(}\text{$p_i$, $p_j$}\textbf{)}\;
    \tcp{\scriptsize radio propagation model}
    Compute $ERP \gets$ \textbf{signal\_propagation(}\text{$dist$}\textbf{)}\;
    \tcp{\scriptsize smoothing function}
    Compute $csi \gets$ \textbf{csi\_smoothig(}\text{$ERP$, $c_{min}$}\textbf{)}\;

    \Return{$csi$}\;
}
\Return{None}\;
\end{algorithm}

With the implemented modular structure and the integration of radio-technology characteristics, the proposed communication model aims to provide a flexible, easy-to-use approach for assessing inter-agent connectivity, suitable for different NART missions and restrictions.

\subsection*{Implementation for SpArC-NARTs}

An individual communication index, $CSI_i$ is assigned to each NART agent. It represents its connectivity value at each time step. All agents can have several possible links at each time, one with each NART member. Thus, multiple pairwise interactions can occur if communication requirements are met. As only direct links (i.e., pairwise interactions) have been considered in the proposed work, the individual communication index of an agent $i$ is defined as:
\begin{equation}\label{CSI}
    CSI_i(s) = \max_{j \in \mathcal{NART} , i \neq j} csi_{ij}(s) .
\end{equation}

This definition represents the best option that agent $i$ has to communicate with another NART agent. If a tightly connected NART was aimed, an average of the communication indexes of the agent $i$ with the respective neighbours would be preferred. Note that the link between two agents will have a communication index while they are both in the mission (i.e., with available battery).

\subsubsection*{Environments with Obstacles}
The proposed communication model also provides an initial approach to reflect the effect of obstacles in inter-agent connectivity. The radio propagation model was considered under conditions where the Friis equation is valid. This equation is a particular case of the Log-Distance Model. The value of the path-loss exponent, $n$, can vary between 2 (e.g., Friis conditions) and 6, as presented in \cite{Rappaport}, and can be updated to represent the propagation environment in which the NART will be deployed, including the effects of buildings. This parameter can also be estimated as in \cite{9148432} and \cite{6981955}.

Further modifications may include adding log-normal shadowing to the Friis propagation model, as presented in \cite{8114199}. Distinct radio propagation models can also replace the one presented in this work (step 2), as long as their inputs and outputs are compatible with the previous and following steps.

\subsection{Cooperation}

The NART trajectories are optimised prior to the start of the mission based on the mission goals and estimated communication availability. The trajectories indirectly account for cooperation opportunities where agents can share data with other NART agents. A data exchange, and therefore a rendezvous, is considered to occur in a single timestep $s$. Pairwise rendezvous are considered, instead of global rendezvous (i.e., rendezvous in which ideally all agents participate).

In a data exchange, each agent shares its individual knowledge of the environment accumulated until $s$. This knowledge is updated in a history structure, $history\_cell$. The accumulated knowledge encompasses the knowledge gained directly by the agent through exploration and the knowledge it has acquired in previous cooperative behaviours.

As an example, after a data exchange with agent $j$, agent $i$ aggregates the received information (e.g $history\_cell[j]$) with its own (e.g., $history\_cell[i]$), according to Algorithm \ref{data_aggregation}.  $\mathcal{VC}$ is considered the set of cells in the environment, which are valid for exploration and $|\mathcal{VC}|$ the number of valid cells in the environment. 

$history\_cell[i][i][c]$ represents the number of times agent $i$ has visited cell $c$. If agent $i$ does not perform an exploration task, this slot has a value of 0. $history\_cell[i][j][c]$ presents the most recent information regarding the number of times that agent $j$ has visited cell $c$, according to the situational awareness of agent $i$. Agent $i$ can receive this information directly from agent $j$ or indirectly from a different NART member who has previously cooperated with agent $j$.

\begin{algorithm}[!htbp]
\caption{Situational Awareness Update of Agent $i$ after a Data Exchange with Agent $j$.}
\label{data_aggregation}

\KwIn{$history\_cell$, $\mathcal{NART}$, $i$, $j$, $\mathcal{VC}$}
\KwOut{$HC[i]$}

\SetAlgoLined

\textbf{Initialize} $HC \gets history\_cell$

\For{$c \in \mathcal{VC}$}{
    \For{$n \in \mathcal{NART}$}{
        $HC[i][n][c] \gets \max(HC[i][n][c], HC[j][n][c])$ 
    }
}

\Return $HC[i]$

\end{algorithm}

At the end of a data exchange, the number of times that cell $c$ has been visited according to the situational awareness of agent $i$ at time step $s$ is given by Eq. \ref{v_ics}:

\begin{equation} \label{v_ics}
    v_{i, c, s}(c,s) = \sum_{n \in \mathcal{NART}} HC[i][n][c]
\end{equation}

This procedure shows the impact of cooperation in increasing the situational awareness of both agents. During the mission, these rendezvous may also be used to exchange relevant mission information, such as the discovery of the targets, individual sensor measurements and mission modifications.

\subsection*{Value of Movement ($VoM$)}

Hard restrictions for reporting such as fixed time intervals between reportings have been explored in the literature to promote inter-agent communication. These, however, often disregard of the impact on the exploration mission. As inter-agent communication often implies spatial proximity between them, it can significantly jeopardise their ability to disperse in the environment and explore it. Its negative impact is particularly significant when deploying a small NART team with limited capabilities. Thus, in this proposal, more flexible reporting times are considered to reduce the impact on exploration goals while still promoting data exchanges. 

The Value of Movement ($VoM$), initially proposed by \cite{139}, is a dynamic time-based incentive for communication. This individual parameter increases while no data exchanges are performed up to a maximum value and decreases to a minimum value once a data exchange is performed. 

In the proposed work, the $VoM$ was adapted to be a time-evolving parameter which simultaneously quantifies the time elapsed between data exchanges of an agent and weights accordingly its urgency to report. The urgency to report increases exponentially with the time elapsed from the most recent data exchange of an agent. Most exploration-based applications do not benefit from highly frequent reporting as it can overly restrict the motion of the agent and reduce overall information gains. In these settings, reporting should also be avoided shortly after a data exchange. Reporting should thus be incentivised (i.e., rewarded) if an agent has not exchanged data with any neighbour for a long period of time and discouraged (ie., penalised) if a data exchange took place recently. These behavioural rewards and penalties act as soft-restrictions for communication. 

The new formulation of the $VoM$ is presented in Eq. \ref{new_VoM} which integrates:
\begin{itemize}
    \item Monotonic time-based increase with the time elapsed since the last successful data exchange.
    \item Reset upon a successful data exchange.
    \item Soft-restrictions for communication as an output: reward ($VoM\geq0$) or penalty ($VoM<0$).
\end{itemize}

\begin{equation}\label{new_VoM}
  VoM_i(s) =
    \begin{cases}
      \frac{2e^{\frac{s-s_{i}}{T_{sys}}}-1}{e^{\frac{\tau}{T_{sys}}}-1}-1 & \text{if $s - s_{i} \leq \tau_i$}\\
      1 & \text{otherwise}
    \end{cases}  .
\end{equation}

In this case, $s_{i}$ represents the step of the last data exchange of agent $i$. The parameter $\tau_i = \frac{lifetime_i}{n_{meetings}}$, represents the maximum interval between meetings. $n_{meetings}$ is a reference (soft-constraint) for the number of meetings that a NART agent has during a mission. This parameter is defined by the user. $lifetime_i$ is the number of steps in the path of $UAV_i$, which is mostly limited by its available battery. A positive value of $VoM$ corresponds to an incentive for communication. A negative value of $VoM$ will signal a need to discourage communication, enabling other mission components (e.g. exploration) to be prioritised.

Individual dynamic $VoM$ evolution can lead to slightly larger or shorter time intervals between meetings than the reference $\tau$. This flexibility supports a balanced integration of inter-agent communication with other mission goals and overall optimised mission performance. If very frequent reporting is aimed, providing a large reference of meetings ($n_{meetings}$) will create more frequent incentives to communication. Such may be important for NART application requiring a high level of global situational awareness regardless of the impact on agent motion. A lower value of $n_{meetings}$ will incentivise sparser meetings and reduced motion restrictions. The frequency at which this parameter is updated is defined by the user with $T_{sys}$. 

This approach significantly improves the flexibility of inter-agent communication, through softly-defined incentives instead of hard constrained rendezvous scheduling.

\subsection*{Dynamic Trade-off between Exploration and Communication}

In this work, cooperation is a communication-dependent event, defined as in Eq. \ref{coop}, that simultaneously considers a need to report and an ability to communicate. The reporting requirement derives from expected mission development and the parameter $n_{meetings}$. The ability to communicate reflects the limitations of user-defined radio technology and the influence of physical phenomena (e.g., signal fading). 
\begin{equation}\label{coop}
    C_i(s) = VoM_i(s) \cdot CSI_i(s).
\end{equation}

\begin{figure}[htbp]
    \centerline{\includegraphics[scale=0.32]{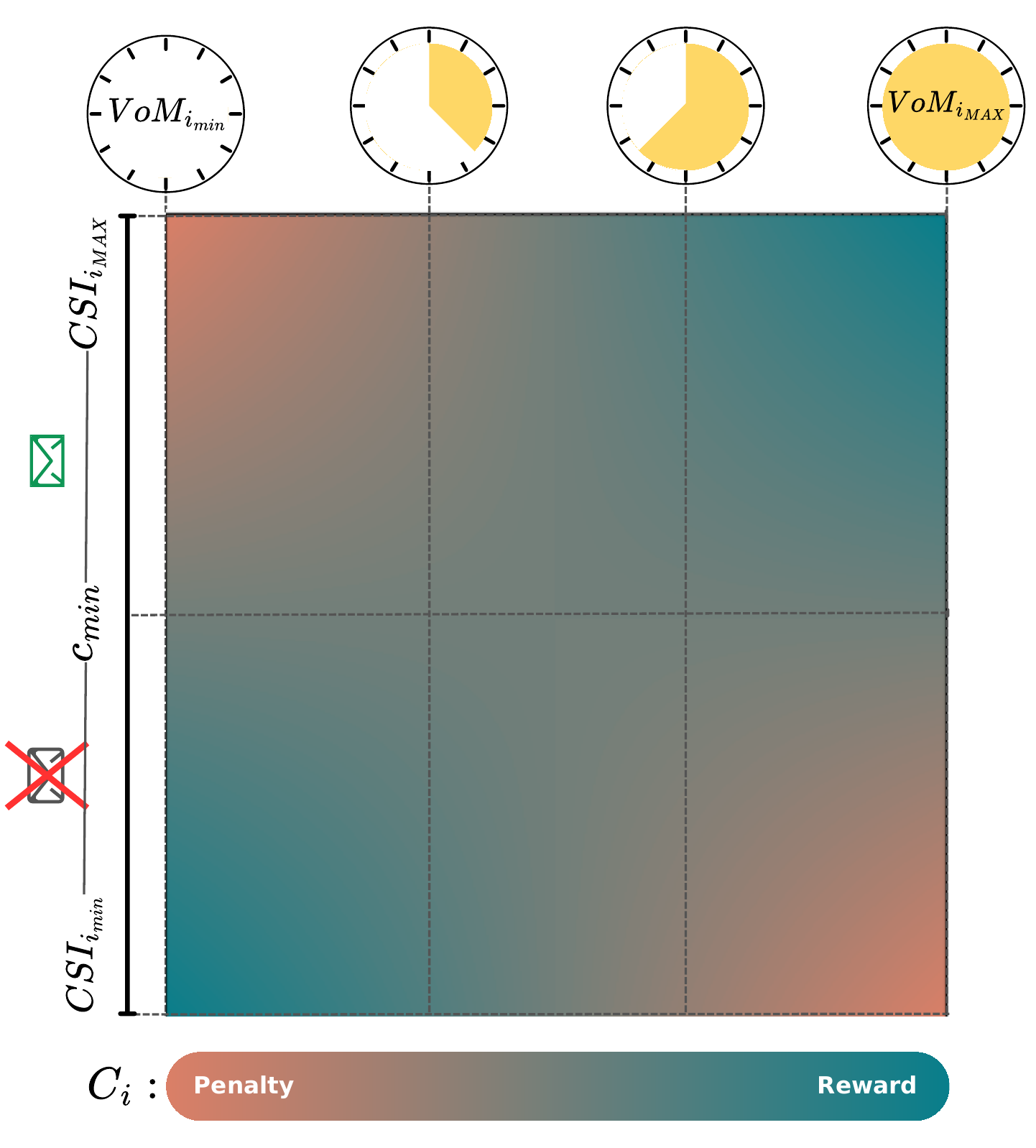}}
    \caption{Exploration-Communication Dynamic  Rewards. At the top is the evolution of $VoM_i$. $VoM_i$ reaches its minimum value right after a data exchange has taken place. On the side, the value ranges of $CSI_i$. A data exchange is possible if $CSI_i \geq c_{min}$.}
    \label{DynamicCoopIncentives}
\end{figure}

The cooperation component $C_i(s)$ comprises two agent components at each step: its need for communication, given by the proposed adaptation of $VoM$, and its respective ability to exchange data, given $CSI_i(s)$. The $C_i(s)$ value represents a dynamic reward or a penalty, as presented in Fig. \ref{DynamicCoopIncentives}.

When an agent requires communication (i.e. has urgency to communicate), $VoM_i > 0$. Given $CSI_i = CSI_j = csi_{i,j}$, a reward is given if the $csi_{i,j}$ between the agent $i$ and a teammate $j$ allows a data exchange between them. Consequently, if the $csi_{i,j}$ is negative, a penalty is given, since communication is not possible even though agent $i$ has an urgency to report. 

\begin{figure}[!t]
    \centerline{\includegraphics[scale=0.4]{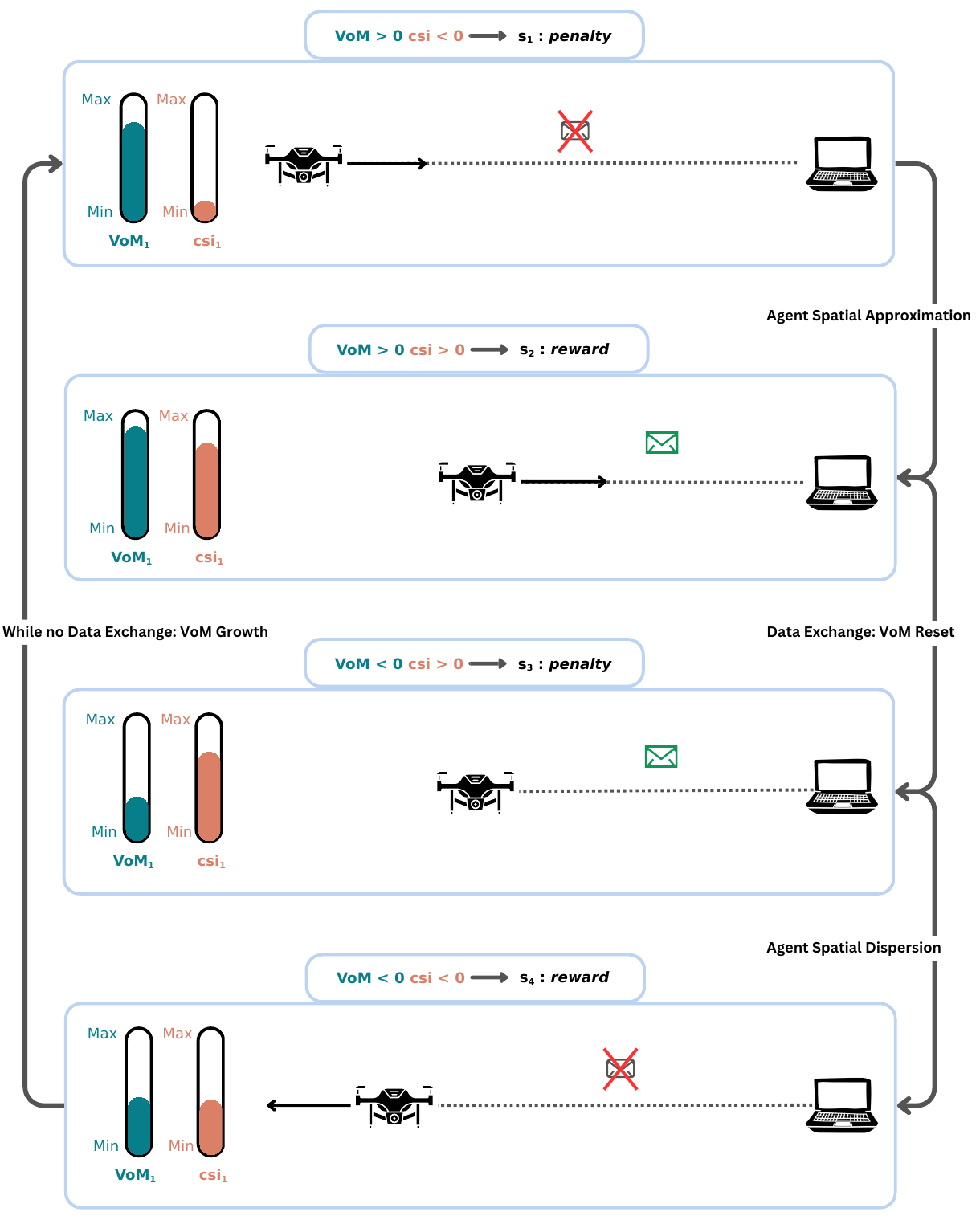}}
    \caption{Exploration-Communication Dynamic  Behavioural Loop. Impact of Behavioural Loop in NART agent motion, for a NART constituted by a UAV and a Ground Station.}
    \label{loop_comm}
\end{figure}

Once a data exchange takes place, the $VoM$ of both agents is reset to the minimum value. The agents must disperse after a data exchange to continue exploring the environment. In this case, the $VoM$ of each agent has a negative value. Each agent will then receive a reward if their $CSI$ is also negative. Otherwise, a penalty is given. The dynamic trade-off between exploration and communication can be explained as a high-level behavioural loop, as illustrated in Fig. \ref{loop_comm}.

\subsection*{Role-based Behaviour Loop}
So far, cooperation considers that each agent has a single behavioural loop and an individual $VoM$ that evolves as the agent interacts with all other NART agents, regardless of their mission role. 

Some applications, however, might benefit from a distinction between cooperating with or reporting to agents with different capabilities (i.e., roles). Reporting to a static external entity (e.g., GCS, command post) may be preferred over reporting to neighbour UAVs, due to fewer limitations on computational power or battery life. EEs can also operate as hubs or relays for UAVs without direct communication.

The formulation of cooperation in this setting accounts for separate $VoM$s for inter-UAV and UAV-EE interactions and weights them differently, as introduced in \cite{139}. Cooperation considering a distinct role-based behaviour loop $C_{i_{RB}}(s)$ can be formulated as:

\begin{equation}\label{coop_sep}
    C_{i_{RB}}(s) = w_1 C_{i_\mathcal{ETD}}(s)+  w_2 C_{i_\mathcal{R}}(s).
\end{equation}

The parameters $w_1$ and $w_2$ weight the inter-UAV and UAV-EE links, respectively. The cooperation between agent $i$ and all explorer agents is represented by $C_{i_\mathcal{ETD}}$. $C_{i_\mathcal{R}}$ represents the cooperation between agent $i$ and all reporting agents in the NART. $C_{i_\mathcal{ETD}}$ and $C_{i_\mathcal{R}}$ follow the formulation presented in Eq. \ref{coop}, with the distinction that instead of considering all NART agents, they only consider the interactions with explorer or relaying agents, respectively.

\subsection{Complexity Analysis}
The complexity bottleneck of the developed work lies in the cost computation of each candidate solution in the Simulated Annealing. A two-step process is repeated for each time step of the mission duration. The outer-loop is $\mathcal{O}(P)$, where $P$ represents $max\_path\_len$. 

The first part assesses the exploration cost of all agents ($\mathcal{O}(A)$), where $A$ represents the number of agents in the NART. The second part analyses the communication and cooperation costs across all links, $L$ (i.e., unique pairs of agents), resulting in $\mathcal{O}(L)$, which can be presented as $\mathcal{O}(A^2)$. The aggregation of the two sequential steps results in $\mathcal{O}(PA^2)$. 

Algorithm \ref{data_aggregation} runs within the second step when the communication index of a given link (e.g. $c_{01}$) is compatible with a data exchange. It has a complexity of $\mathcal{O}(VA)$. $V$ represents the number of valid cells in the environment ($\mathcal{VC}$). The proposed work considers the deployment of sparsely connected agents. Algorithm \ref{data_aggregation} will thus be applied in a fraction ($f \leq 1$) of the steps, resulting in $\mathcal{O}(fVA)$. 

The overall complexity of the cost computation function is $\mathcal{O}(PA^3V)$ in the worst case ($f=1$). However, it is important to note that the SpArC-NART is considered to be deployed in an environment which greatly exceeds its individual capabilities ($V \gg A^3$).

\section{SpArC-NARTs Optimisation for a Ground Sensing Coverage Mission} \label{CPPUseCase}

The Ground Sensing Coverage Mission was chosen as a use case. Planned and opportunistic cooperation are considered as communication-dependent events that operate on user-selected communication technologies. Agents may act as explorers and/or relays during the mission. Roles are not hard-coded but encouraged by soft constraints and rewards. The developed software communication model computes a pairwise communication strength index based on agent poses and a user-selected technology. It considers specific restrictions of the selected technology, such as signal fading and packet exchange reliability (through the sensitivity of the receiver). The reports allow the exchange of important mission updates or information collected during the development of the optimised trajectories.
Additionally, if required, the reporting between agents can also be used as an opportunity for online mission replanning. The weight given to incentives for cooperation opportunities and the entities to report are both user-selected parameters. Different combinations of parameters reflect different high-level behaviours of the agents. For this reason, this offline optimisation can be applied in several NART applications. 
$\mathcal{ETD}$ refers to the set of explorer agents (e.g., agents assigned to the task of exploration and target detection) and $\mathcal{R}$ refers to the set of relaying agents (e.g., agents assigned to the task of reporting).

\subsection{NART Agent Kinematics for Ground Sensing Coverage Use Case}

In this work, a $\Delta z_{agent}$ of 2[m] is considered for the altitude displacement between agents. The trajectories of the SpArC-NART agents are optimised resorting to the Simulated Annealing (SA) Algorithm. SA is a metaheuristic with classical asymptotic convergence to a global optimum under sufficiently slow cooling schedules. Multi-SA-threading and Markov chains, similar to \cite{OriginalPP}, as were randomly generated initial configurations, as suggested in \cite{Kirkpatrick}, were considered. These strategies have been implemented in different applications such as \cite{LEE2011707}, \cite{ferreiro2013efficient}.

The initial positions of the UAVs are included in the optimisation. The positioning and energy limitations of the EEs were not included in the optimisation, and the EEs are considered to have sufficient energy to support the complete NART mission. The path of M-EE is pre-planned according to Alg \ref{M-EE_path}. Firstly, the Dijkstra Algorithm is applied between the vertices of the area of interest (AoI), along adjacent cells close to the frontier of the area. With the weighted graph G, the path with the shortest distance between two AoI vertices is found. 

The purpose of the integration of an M-EE is to support the UAVs during their full mission duration (e.g., $max\_path\_len$). Therefore, the length of the M-EE trajectory (i.e., the duration of the M-EE mission) was considered to be equal to the length of the longest UAV trajectory (i.e., the UAV with the longest mission duration). The default duration of the M-EE mission, considering a trajectory that covers the perimeter of the environment a single time, can be shorter or longer than $max\_path\_len$. Therefore, the M-EE trajectory can be clipped or extended to equal the UAV with the largest lifetime.

\begin{algorithm}[t] 
\caption{Mobile External Entity Path: Generate and Adapt to Mission Duration.}
\label{M-EE_path}
\SetAlgoLined
\KwIn{$AoI\_vertices$, $max\_path\_len$, $\mathcal{VC}$, $adj\_list$}
\KwOut{$M\_EE\_path$}

Initialize $base\_path \gets \emptyset$ \;

Initialize $M\_EE\_path \gets \emptyset$ \;

G $\gets$ \textbf{weighted\_graph(}$\mathcal{VC}$\textbf{)}\;
\tcp{\scriptsize find shortest path along AOI frontiers}
\For{vertex $v$ in range($AoI\_Vertices$)}{
    \If{$v$ is the last vertex in $AoI\_Vertices$}{
        $next\_v$ $\gets$ 0\;
    }
    \Else{
        $next\_v$ $\gets$ $v+1$\;
    }
    Compute $p \gets$ \textbf{Dijkstra(}\text{$v$, \text{$next\_v$}, $G$}\textbf{)}\;
    Extend $base\_path$ with $p$\;
}
\tcp{\scriptsize reconstruct: list of indexes to list of cells}
$M\_EE\_path \gets$ \textbf{reconstruct\_path(}\text{$base\_path$, $\mathcal{VC}$}\textbf{)}\;

\tcp{\scriptsize adapt $M\_EE\_path$ to mission length}
$M\_EE\_path \gets$ \textbf{adapt\_path(}\text{$M\_EE\_path$, $max\_path\_len$}\textbf{)}\;

\Return{$M\_EE\_path$}\;

\end{algorithm}

\subsection{Use Case Definition}

The use cases may consider non-cooperative strategies (No Coop) or a cooperative strategy. Opportunistic cooperation (OCoop) can occur in all the use cases, whenever communication requirements are met. The planned cooperation strategy can derive from the optimized trajectories with a dynamic reward-based balance between exploration and reporting (PCoop-DR) or with a global rendezvous approach with inter-meeting fixed intervals and inter-agent-based connectivity (PCoop-GR). 

Three mission strategies were compared: a non-cooperative baseline, a cooperative baseline and the proposed strategy. All missions strategies consider the same prior knowledge and energy limitations, but distinct cooperation strategies and requirements for connectivity.

\subsection*{Baselines}

The first baseline, similar to \cite{OriginalPP}, does not explicitly consider the cooperation of NART agents. At most, opportunistic cooperation may occur only if the paths of the agents happen to overlap under the communication range. This strategy will be considered as a non-cooperative baseline approach (Baseline A). The communication requirement in this strategy follows the proposed communication model. 
\vspace{2mm}

Two cooperative baselines (Baselines B and C) were considered. These explicitly considers the cooperation of NART agents, under the following strategies:
\begin{itemize}
\item A rendezvous is considered a cluster (or global) event, similar to \cite{LUPERTO2025105137} and \cite{10679913}.
\item Fixed time intervals between rendezvous, similarly to the exploring time in \cite{9981898}.
\end{itemize}

\vspace{2mm}
Furthermore, in Baseline B, the communication requirement does not account for radio-technology characteristics, in this case being the expected communication range that is shorter than the one obtained by a more realistic propagation model. It considers an inter-agent euclidean distance, similar to \cite{124}, \cite{10679913}. A communication range of 10 [m] was considered. Alternatively, in Baseline C, the proposed communication model is considered. In both baselines, the meeting points are defined indirectly and result from optimising the NART trajectories under the aforementioned cooperation and connectivity constraints.

Baseline A will allow the evaluation of the impact of cooperation on mission performance. Baseline B will enable comparison of the proposed dynamic incentives with the cooperation and communication model relative to other cooperation strategies and connectivity requirements in the literature. Baseline C will enable the evaluation of the impact of higher communication awareness, particularly the limitations of user-defined radio technology, relative to frequent cooperation strategies in the literature.

\subsection*{NART Settings}

A team of three UAVs was considered under four use cases with different NART settings. The multi-UAV group (MUG) use case tests the capabilities of an independent multi-UAV group without the support of an external entity. The static external entity (S-EE) use case considers a multi-UAV group supported by an entity such as a GCS. The M-EE use cases consider a multi-UAV group supported by a mobile external entity (e.g., a larger aircraft, a dedicated UAV, or a data mule). 

Two M-EE use cases were defined to assess the effect of role-based behaviour loop in NART performance. The first one, M-EE1, considers that the behaviour of each agent reflects its interactions with all NART agents. In the second one, M-EE2, each agent considers role-based distinction with dedicated $VoM$s for inter-UAV and UAV-EE interactions.

The use cases also account for direct and indirect reporting to external entities. While the S-EE use case account only for direct reporting to the external entity (e.g., GCS), the M-EE use cases consider both direct and indirect reporting between all agents.

\begin{table*}[!htp] 
\centering
\caption{Use Case Definition.}
\label{use_cases}
\resizebox{\textwidth}{!}{
\begin{tabular}{cc|ccccc|cc|c|}
\cline{3-10}
\multicolumn{1}{l}{} & \multicolumn{1}{l|}{} & \multicolumn{5}{c|}{\textbf{Cooperation}} & \multicolumn{2}{c|}{\textbf{Task}} & \multicolumn{1}{c|}{\multirow{2}{*}{\textbf{Connectivity}}} \\ \cline{3-9}
\multicolumn{1}{l}{} & \multicolumn{1}{l|}{} & \multicolumn{1}{c|}{Strategy} & \multicolumn{1}{c|}{$w_1$} & \multicolumn{1}{c|}{$w_2$} & \multicolumn{1}{c|}{Role-Based Behaviour} & Reporting Times & \multicolumn{1}{c|}{$ETD$} & R & \multicolumn{1}{c|}{} \\ \hline
\multicolumn{2}{|c|}{Baseline A, \cite{OriginalPP}} & \multicolumn{1}{c|}{No Coop, OCoop} & \multicolumn{1}{c|}{0} & \multicolumn{1}{c|}{0} & \multicolumn{1}{c|}{No} & - & \multicolumn{1}{c|}{UAVs} & - & Proposed CM \\ \hline
\multicolumn{2}{|c|}{Baseline B} & \multicolumn{1}{c|}{PCoop-GR, OCoop} & \multicolumn{1}{c|}{0} & \multicolumn{1}{c|}{0} & \multicolumn{1}{c|}{No} &Fixed Reporting Times & \multicolumn{1}{c|}{UAVs} & - & Euclidean Distance \\ \hline
\multicolumn{2}{|c|}{Baseline C} & \multicolumn{1}{c|}{PCoop-GR, OCoop} & \multicolumn{1}{c|}{0} & \multicolumn{1}{c|}{0} & \multicolumn{1}{c|}{No} & Fixed Reporting Times & \multicolumn{1}{c|}{UAVs} & - & Proposed CM \\ \hline
\multicolumn{1}{|c|}{Without External Support} & \multicolumn{1}{|c|}{MUG Use Case} & \multicolumn{1}{c|}{PCoop-DR, OCoop} & \multicolumn{1}{c|}{1} & \multicolumn{1}{c|}{0} & \multicolumn{1}{c|}{No} & Proposed $VoM$& \multicolumn{1}{c|}{UAVs} & UAVs & Proposed CM \\ \hline
\multicolumn{1}{|c|}{\multirow{3}{*}{With External Support}} & S-EE Use Case & \multicolumn{1}{c|}{PCoop-DR, OCoop} & \multicolumn{1}{c|}{0} & \multicolumn{1}{c|}{1} & \multicolumn{1}{c|}{No} & Proposed $VoM$ & \multicolumn{1}{c|}{UAVs} & EE & Proposed CM \\ \cline{2-10}
\multicolumn{1}{|c|}{} & M-EE1 Use Case & \multicolumn{1}{c|}{PCoop-DR, OCoop} & \multicolumn{1}{c|}{0.3} & \multicolumn{1}{c|}{0.7} & \multicolumn{1}{c|}{No} & Proposed $VoM$ & \multicolumn{1}{c|}{UAVs} & UAVs, EE & Proposed CM \\ \cline{2-10}
\multicolumn{1}{|c|}{} & M-EE2 Use Case & \multicolumn{1}{c|}{PCoop-DR, OCoop} & \multicolumn{1}{c|}{0.3} & \multicolumn{1}{c|}{0.7} & \multicolumn{1}{c|}{Yes} & Proposed $VoM$ & \multicolumn{1}{c|}{UAVs} & UAVs, EE & Proposed CM \\ \hline
\end{tabular}
}
\vspace{1mm}
\noindent\footnotesize\textit{Note:} Communication Model (CM).
\end{table*}

\subsection*{Common Parameters} 

The use cases defined consider different task distributions among NART agents. Each NART agent can be considered part of the set of agents assigned to the Exploration ($\mathcal{EDT}$) or the Reporting ($\mathcal{R}$) Tasks. In the S-EE use case, the UAVs and the EE are respectively assigned to the exploration and reporting tasks. In the M-EE use cases, the EE always takes the reporting task. In the MUG and M-EE use cases, the UAVs are assigned to both tasks.

All UAVs are considered to have 2000 units of energy. The energy consumption of the UAVs is maintained as in \cite{OriginalPP}. It considers the energy required for the translation and rotation movements that the agent performs at each step $s$ between adjacent cells in the environment. Considering, as an example, a UAV trajectory without rotations, 2000 units of energy would be sufficient for a mission duration of around 86 timesteps. It was also considered a probability of detection ($pod$) of 63\%. 

Regarding the user-defined $VoM$ parameters, a reference of 4 reports per agent ($n_{meetings}$) was considered. The time interval between rendezvous in Baseline B was also defined to achieve a similar reference of 4 global meetings. All cooperative strategies considered a step-wise reward update (i.e. $T_{sys} = 1$). 5 trials were considered for each approach (e.g., baselines and use cases). A single set of initial S-EE positions was tested in both environments, one for each trial, including the centre and edges of the environment. The main parameters are shown in Table \ref{use_cases}.

The parameters considered in the communication model are presented in Table \ref{Comm_Params}, corresponding to an estimated communication range of approximately 445 [m]. As the user provides these parameters, different radio technologies can be tested.

\begin{figure}[!thpb] 
    \centering
    \begin{subfigure}[b]{\columnwidth} %
      \centering
      \includegraphics[width=0.9\linewidth]{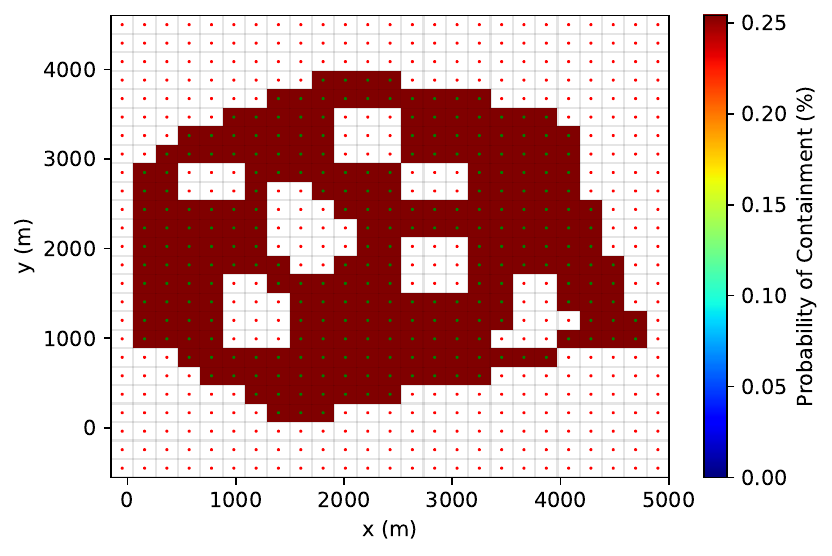}
      \caption{Uniform POC Distribution.}
      \label{mission_env_a}
    \end{subfigure}
    \hfill
    \begin{subfigure}[b]{\columnwidth}
      \centering
      \includegraphics[width=0.9\linewidth]{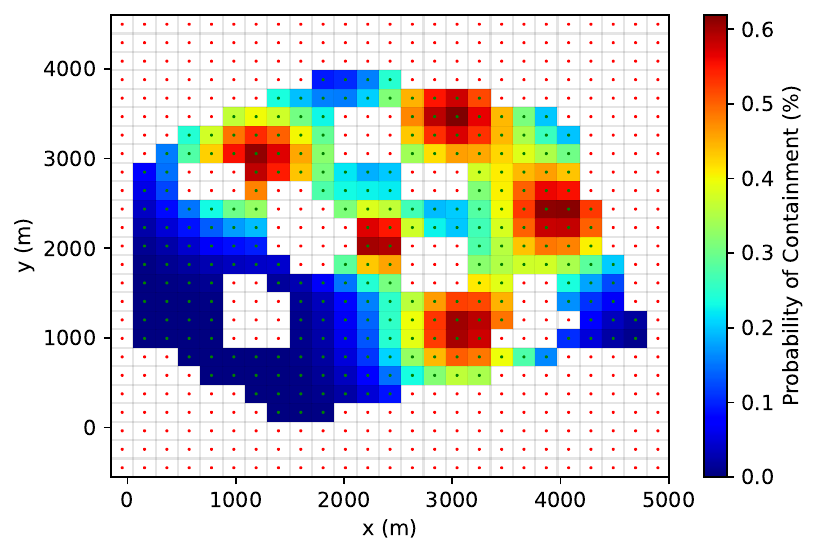}
      \caption{Non-Uniform POC Distribution.}
      \label{mission_env_b}
    \end{subfigure}
    \caption{Mission Environments with distinct POC distributions. Total POC = 64.80\%.}
\end{figure}

\begin{table}[!htbp] 
\centering
\caption{Communication Parameters used in the Use Cases.}
\label{Comm_Params}
\resizebox{0.4\textwidth}{!}{ 
\begin{tabular}{c|c|c|}
\cline{2-3}
\multicolumn{1}{l|}{}                                          & \textbf{Parameter}  & \textbf{Value}                    \\ \hline
\multicolumn{1}{|c|}{\multirow{2}{*}{\textbf{CSI Smoothing}}}  & $k$         & 0.4             \\ \cline{2-3}
\multicolumn{1}{|c|}{}                                         & $\epsilon$  & $1E-6$                      \\ \hline
\multicolumn{1}{|c|}{\multirow{3}{*}{\textbf{Radio Technology}}}        & Protocol   & IEEE 802.11g             \\ \cline{2-3} 
\multicolumn{1}{|c|}{}                                         & $Pt$       & 0.100 [W]                  \\ \cline{2-3} 
\multicolumn{1}{|c|}{}                                         & $c_{min}$  & -73 [dBm]                      \\ \hline
\multicolumn{1}{|c|}{\multirow{6}{*}{\textbf{Radio Propagation Model}}} & Name       & Friis Propagation \\ \cline{2-3} 
\multicolumn{1}{|c|}{}                                         & $G_t$       & 1                        \\ \cline{2-3} 
\multicolumn{1}{|c|}{}                                         & $G_r$       & 1                        \\ \cline{2-3} 
\multicolumn{1}{|c|}{}                                         & $c$          & $3E8$  [m/s]                     \\ \cline{2-3} 
\multicolumn{1}{|c|}{}                                         & $f$          & $2.4E9$ [Hz]                   \\ \cline{2-3} 
\multicolumn{1}{|c|}{}                                         & $n$          & 2                        \\ \hline
\end{tabular}
}
\end{table}
\vspace{1mm}
Two mission environments were considered, which differ in the prior knowledge of the environment. Uniform and non-uniform dispersion of the probability of finding targets was considered. The resulting probability of containment for each cell is represented in the environment maps in Fig. \ref{mission_env_a} and Fig. \ref{mission_env_b}, respectively. The coordinate (x,y) = (0,0) is therefore the origin of the inertial frame $\{I\}$, a static global frame.

\pagebreak

\subsection{Optimisation Objectives}

The Ground Sensing Coverage use cases consider three goals:

\begin{itemize}
    \item Goal 1: Maximize Probability of Finding Targets.
    \item Goal 2: Reduce Reporting Latency.
    \item Goal 3: Improve Agent Global Situational Awareness.
\end{itemize}

Maximising the probability of finding targets (Goal 1) was considered the primary goal of the mission. Goal 1 is also the primary objective of the Baseline A and Baseline B. Goals 2 and 3 are considered secondary goals. These reflect the integration of cooperation in the NART mission and its impact on reporting capabilities and situational awareness of explorer agents.

This work considered that the optimiser optimises Goal 1 with a communication-aware shaping term. Goals 2 and 3 are used solely for evaluation purposes. This way, it is possible to evaluate how the introduction of inter-agent cooperation under intermittent connectivity can impact the Ground Sensing Coverage mission performance (e.g., Goal 1), as well as the NART resilience in dynamic or partially known environments (e.g., Goals 2 and 3). 
The objective function considered is presented in Eq. \ref{J}. When the role-based behaviour loop is accounted for, $C_{i_{RB}}(s)$ is considered instead of $C_i(s)$.

\begin{equation}\label{J}
    J = \sum_{s} e^{-\epsilon s} POC_s(s)(1+ \sum_{i \in \mathcal{NART}}C_i(s)) .
\end{equation}

The SA algorithm considered an initial ($T\_init$) and final temperature ($T\_end$) of $1.83E-3$ and $2.11E-5$, respectively, a cooling factor of 0.954 and 15 threads. The initial solution is given by the Attraction Algorithm, as defined in \cite{OriginalPP}. Conditional SA acceptance also follows those defined in the same work.

\subsection{Evaluation Metrics}

Goal 1 is evaluated with the metrics $E$, in Eq. \ref{E}, $TPOC$ in Eq. \ref{toc} and $EP$ in Eq. \ref{EP}. 

\begin{equation} \label{toc}
    \begin{aligned}
    TPOC &= \sum_{s}POC_s(s)\\ 
    &=\sum_{s}\sum_{i \in \mathcal{ETD}} POC(X_i(s))[1-I_{V_s}(X_i(s))] .
    \end{aligned} 
\end{equation}

\begin{equation}\label{E}
    E(X) = \sum_{s} e^{-\epsilon s} POC_s(s) .
\end{equation}

\begin{equation}\label{EP}
    EP = \frac{\sum_{c \in \mathcal{VC}} I_{V_s}(c)}{|\mathcal{VC}|}.
\end{equation}

$E$ and $TPOC$ metrics were maintained from \cite{OriginalPP}. The metric $TPOC$ measures the probabilities of containment associated with all cells visited by all the explorer UAVs during a mission. The metric $E$ accumulates the probabilities of containment associated with all cells visited by all the explorer UAVs during a mission, weighted by a time factor. Cells visited sooner in the mission have a higher impact on the metric performance. Thus, visiting areas of high interest is prioritised over the remainder. $I_{V_s}(c)$ has a value of 1 if cell $c$ was visited by any explorer agent.

To complement the above metrics, the percentage of explored cells by the NART was also analysed. If all cells in the environment are covered at least once, the exploration percentage ($EP$) would be 100\%. Although this analysis does not account for the probability of containment or sensor uncertainty, which are crucial for the chosen use case, it provides additional insights into the impact that integrating cooperation strategies can have on the ability of agents to explore the environment.

Goal 2 is evaluated through the expected total amount of reports, $ETR$, Eq. \ref{etr}, and the expected average reporting time, $EART$, Eq. \ref{eart}, of the explorer agents.
\begin{equation}\label{etr}
    ETR = \sum_s \sum_{i \in \mathcal{EDT}} Rp_i(s).
\end{equation}
where
\begin{equation}\label{r_i}
    Rp_i(s) = 
    \begin{cases}
    Rp_i(s) +1, &   \text{if } csi_{ij}(s) \ge 0, \forall j \in \mathcal{NART}, i\neq j\\
    Rp_i(s), & \text{if } csi_{ij}(s) < 0, \forall j \in \mathcal{NART}, i\neq j\\
    0, & \text{otherwise}
    \end{cases} .
\end{equation}

\begin{equation}\label{eart}
    EART = \frac{\sum_{i \in \mathcal{EDT}} \frac{Rp_i(lifetime_i)}{lifetime_i}}{|\mathcal{EDT}|}.
\end{equation}
In a data exchange, it is considered that both agents report to each other, resulting in a total of two reports. The $EART$ represents the average time between meetings of the explorer agents during the mission. This metric accounts for the reporting between all agents. 

Goal 3 is evaluated through two metrics that reflect the effect of data exchanges in the individual situational awareness and overall NART situational awareness under intermittent connectivity. The  first accounts for the expected total averaged accumulated knowledge of the NART, $ETAK$, as presented in Eq. \ref{etak}. 

\begin{equation} \label{etak}
    ETAK =\frac{\sum_{s}\sum_{i \in \mathcal{NART}} EAK_i(s)}{|\mathcal{NART}|} ,
\end{equation}
where:
\begin{equation}
    EAK_i(s) = \sum_{c \in \mathcal{VC}} \frac{ P_{{\text{detect}}_i}(c, s)}{|\mathcal{VC}|} ,
\end{equation}
and
\begin{equation} \label{cell_k}
    P_{{\text{detect}}_i}(c,s) = 1-(1-pod)^{v_{i, c, s}} .
\end{equation}

It aggregates the  knowledge that each agent has regarding the cells in the environment. $P_{detect_i}(c, s)$ represents the probability of an agent detecting targets in each cell, given that a target is present in that cell. The evolution of this metric during the mission reflects the individual situational awareness of the agents. It increases through the individual exploration of agent $i$ and the cell knowledge received through cooperation with other NART agents. The parameter $v_{i, c, s}$ represents the number of times the cell $c$ was visited up to step $s$, according to the individual situational awareness of agent $i$.

The second metric addresses the expected intersected knowledge between NART agents, $EIK$, as presented in Eq. \ref{sgsa}.

\begin{equation} \label{sgsa}
    EIK =\frac{\sum_{s}\sum_{c \in \mathcal{VC}} K_{min}(c,s)}{|\mathcal{VC}|},
\end{equation}
where:
\begin{equation} 
    K_{min}(c,s) = \min_{i=1,\dots,|\mathcal{NART}|} P_{{\text{detect}}_i}(c,s),
\end{equation}

The  $EIK$ metric complements $EAK$, as it measures the intersection between the individual situational awareness of NART agents. This intersection is analysed in a cell-based approach, represented through the minimum $P_{{\text{detect}}_i}(c,s)$ value among all NART agents. This metric reflects the effect of individual exploration, but mostly the ability that NART agents have to share their individual knowledge regarding the environment. A non-cooperative NART, without any direct or indirect reports between agents, will achieve a $EIK$ value of 0 when agents explore disjoint regions.

\section{Simulation Results} \label{sim_Results}

The simulation results are presented according to the mission goals. Subsection \ref{goal12} will address the SpArC-NART performance for Goal 1. Subsection \ref{goal3} addresses Goal 2, and Subsection \ref{goal4} focuses on the performance of the use cases related to Goal 3.

\subsection{Exploration and Target Detection} \label{goal12}

\begin{figure}[htp]
    \centering
    \includegraphics[scale = 0.52]{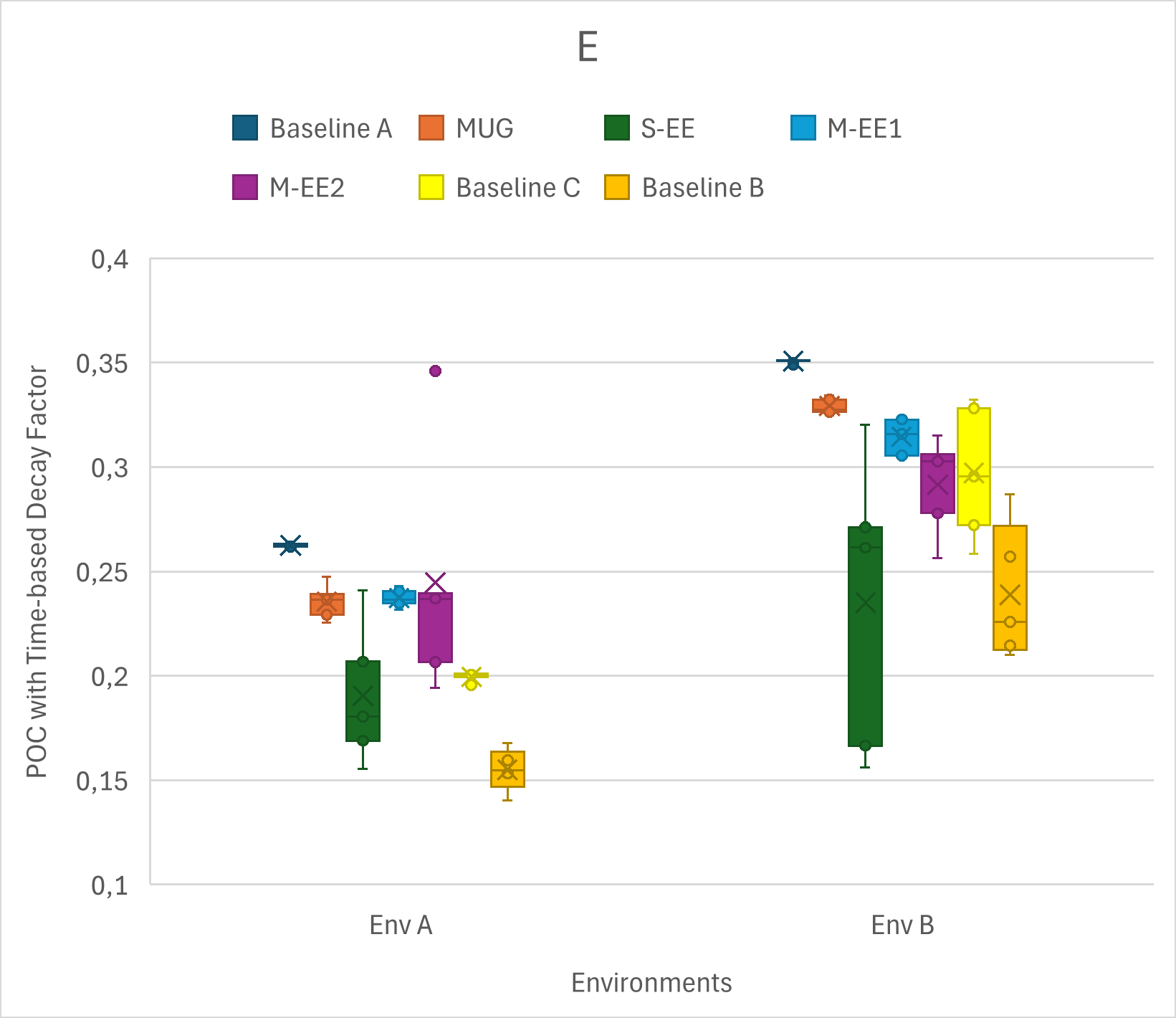}
    \caption{Results for metric $E$.}
    \label{E_metric}
\end{figure}

\begin{figure}[!h]
    \centering
    \includegraphics[scale = 0.52]{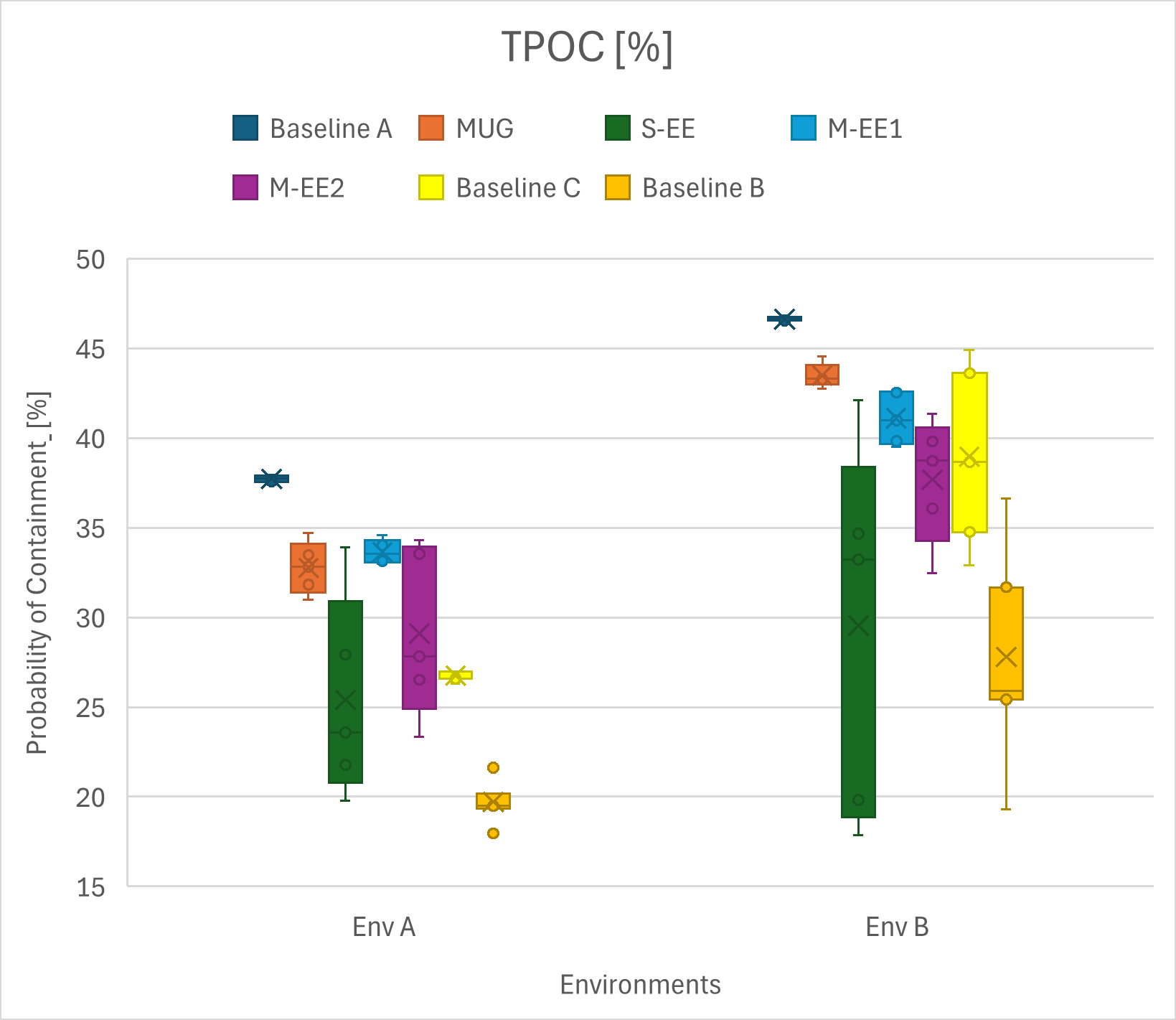}
    \caption{Results for metric $TPOC$.}
    \label{TPOC_metric}
\end{figure}

\begin{figure}[htp]
    \centering
    \includegraphics[scale = 0.6]{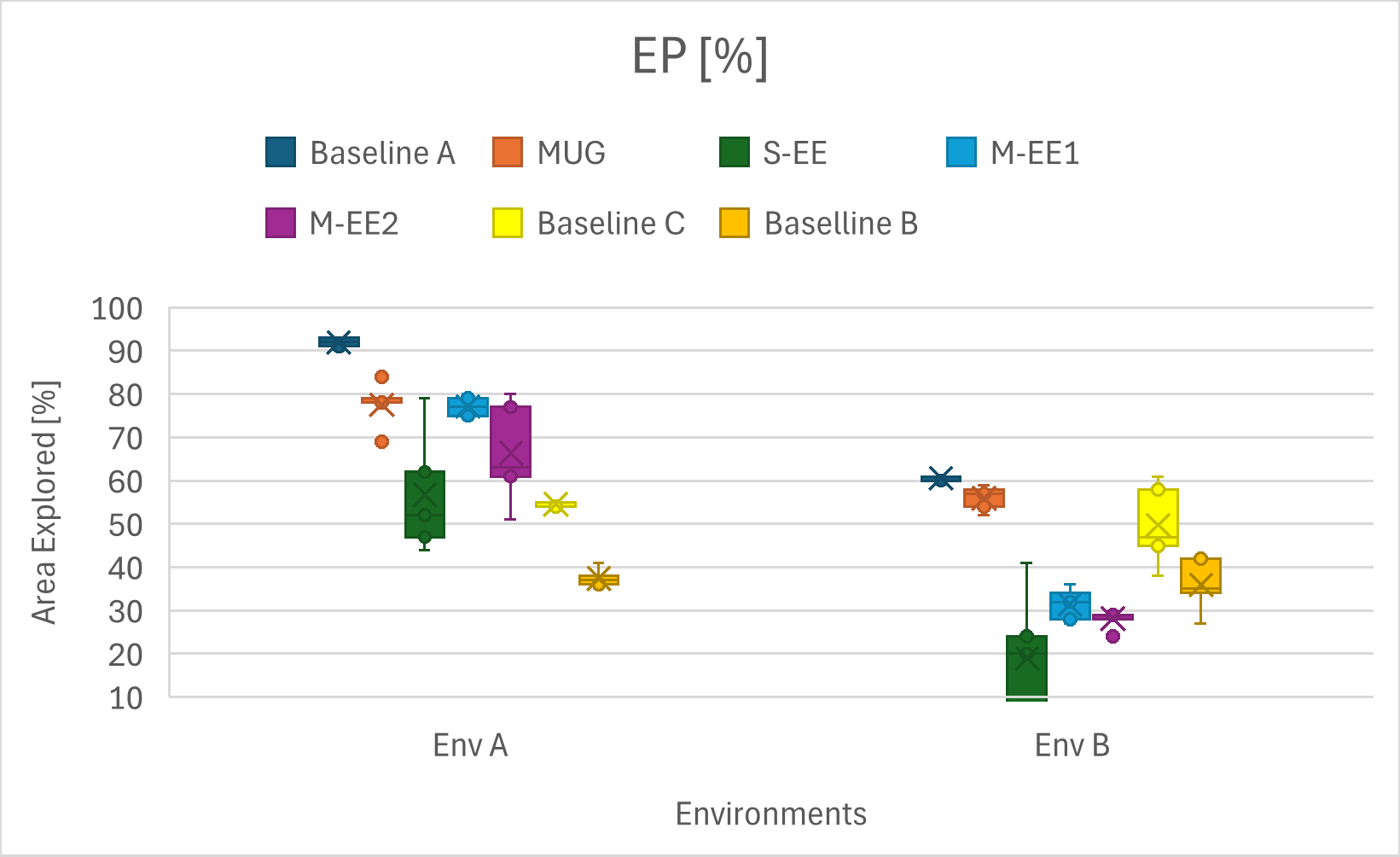}
    \caption{Results for metric $EP$.}
    \label{EP_metric}
\end{figure}

The results obtained for the metrics $E$, $TPOC$ and $EP$ are respectively illustrated in Fig. \ref{E_metric}, Fig. \ref{TPOC_metric} and \ref{EP_metric}, demonstrating overall equivalent behaviours. The addition of cooperation reduced the ability of the exploring agents to cover high-interest cells under the SpArC-NART limitations, when compared to the Baseline A approach. As a data exchange requires agents to be close enough to meet communication requirements according to the selected radio technology, it results in a restriction on agent motion. While the proposed cooperative strategies balance this restriction with the exploration task, the Baseline A focuses solely on dispersing the agents in the environment according to the POC map. 

The rigid constraints on cooperation implemented in Baselines B and C reduced the ability to explore the environment.

The MUG use case had a lower performance than the Baseline A across all metrics related to Goal 1. However, the definition of areas of interest in Env B yielded higher performance and a smaller performance gap than the Baseline A. The performance gap between the Baseline A and MUG use case in Env A was 9.9\% for metric $E$, 13\% for metric $TPOC$ and 15.2\% for metric $EP$, considering average values for each use case. This gap reduced in Env B to 6.7\%, 7.12\% and 6.55\%, respectively, for metrics $E$, $TPOC$ and $EP$. The S-EE use case and Baseline C demonstrated a similar behaviour, with increased performance from Env A to Env B. The performance gaps of the S-EE use case relative to the Baseline A, however, exceed 25\%. Regarding metrics $E$ and $TPOC$ respectively, Baseline C underperformed Baseline A by 23.61\% and 28.53\%  (Env A) and by 15.79\% and 17.01\% (Env B). It has also underperformed the MUG use case by 15.22\%, 17.82\% and 29.49\% in Env A, while in Env B the performance gaps reduced to 9.77\%, 10.65\% and 36.84\%, for metrics $E$, $TPOC$ and $EP$, respectively.

Env B encourages agents to explore smaller sub-areas of the environment due to the integration of prior knowledge regarding the environment. This resulted in less dispersed trajectories for all use cases. Nevertheless, the introduction of cooperation incentives further adapts the trajectories so that agents can be close enough to support intermittent connectivity while still allowing exploration of key areas. Without the support of external entities, there are also no additional incentives to disperse the trajectories of the explorer agents, as there is no associated gain. Brief meetings thus occur between segments of exploration, enabling agents to exchange individual knowledge with a reduced impact on the exploration. The trajectories in the MUG use case balance cooperation opportunities and exploration with spatially condensed yet intertwined trajectories. These reasons support the reduced performance gap in the exploration metrics between the MUG use case and the Baseline A.

Considering a role-based behaviour loop in the NART cooperation led to one of the worse performances on Goal 1. M-EE1 overall maintained the performance gap relative to the Baseline A, with performance gaps of approximately 9.9\%, 12\%, and 16.3\% for metrics $E$, $TPOC$, and $EP$. The M-EE2 use case verified a different evolution of metrics between environments. Overall, the performance gap of M-EE2 is greater than that of M-EE1 in both environments, compared to the Baseline A. Metrics $TPOC$ and $EP$ reduced the performance gap towards the Baseline A from 26.3\% and 31.5\% in Env A to 16.9\% and 26.2\% in Env B, respectively. 

In contrast, the performance gap of M-EE1 for metric $E$ increased from 9.7\% (Env A) to 13.7\% (Env B), with regard to the Baseline A. Unlike metrics $TPOC$ and $EP$, metric $E$ accounts for time discounts applied to the POC values of the agents. Considering role-based incentives increases the gain from reporting in path optimisation. Thus, the path required to cooperate with other explorer agents and the mobile external entity has displaced them from covering more cells in areas of the highest interest sooner in the mission.

Baseline B verified a larger performance gap relative to the non-cooperative Baseline for Env A (12.2\% for $E$, 17.6\% for $TPOC$ and 21.3\% for $EP$). In Env B, the respective gaps were 11.2\%, 12.9\% and 16.4\%. The condensed trajectories resulting from the probability distribution of Env B led to a smaller cost associated with the global rendezvous and euclidean-distance connectivity requirement. In this setting, however, abstracting the capabilities of radio technology in Baseline B severely restricted the exploration performance of the NART in both environments. In Env A, Baseline B obtained worse results than Baseline C by 22.70\% in metric $E$, 26.88\% in metric $TPOC$ and 32\% in metric $EP$. In Env B, Baseline B underperformed Baseline C by 23.57\%, 28.15\% and 18.89\% in metrics $E$, $TPOC$ and $EP$, respectively.

Furthermore, the initial position of the static external entity became a deciding factor for the ability of the NART to explore the environment when areas of interest were defined in Env B. Specific external entity locations in the S-EE use case resulted in equivalent behaviours to other high performing cooperative tests. Placing the EE in the centre of an area of high interest encouraged the UAVs to explore and report in the same part of the environment, thereby encouraging both coverage and reporting simultaneously. Conversely, reporting exclusively to a static entity resulted in an increased UAV motion restriction when the entity is located in an area of low probability of containing a target.

\subsection{Agent Reporting} \label{goal3}

The ability of the NART to report is analysed through the $ETR$ and $EART$ metrics. While the Baseline A approach does not account for cooperation, opportunistic data exchanges may occur if the link between two agents meets the communication requirements, which is reflected in both reporting metrics. The cooperative approaches tested in S-EE, MUG, and both M-EE use cases account for both planned and opportunistic cooperation.

\begin{figure}[htp]
    \centering
    \includegraphics[scale = 0.6]{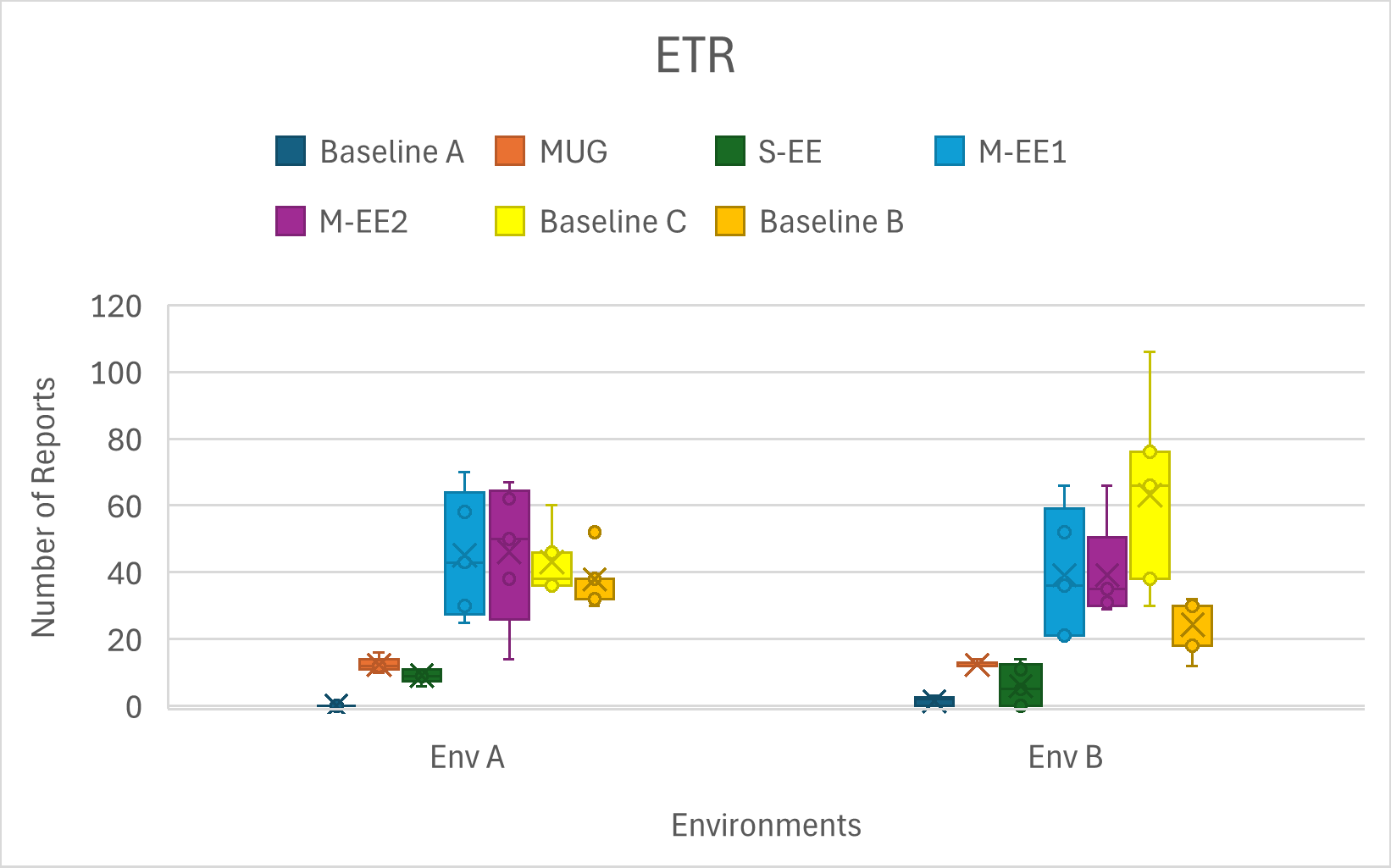}
    \caption{Results for metric $ETR$.}
    \label{ETR_metric}
\end{figure}

\begin{figure}[t]
    \centering
    \includegraphics[scale = 0.6]{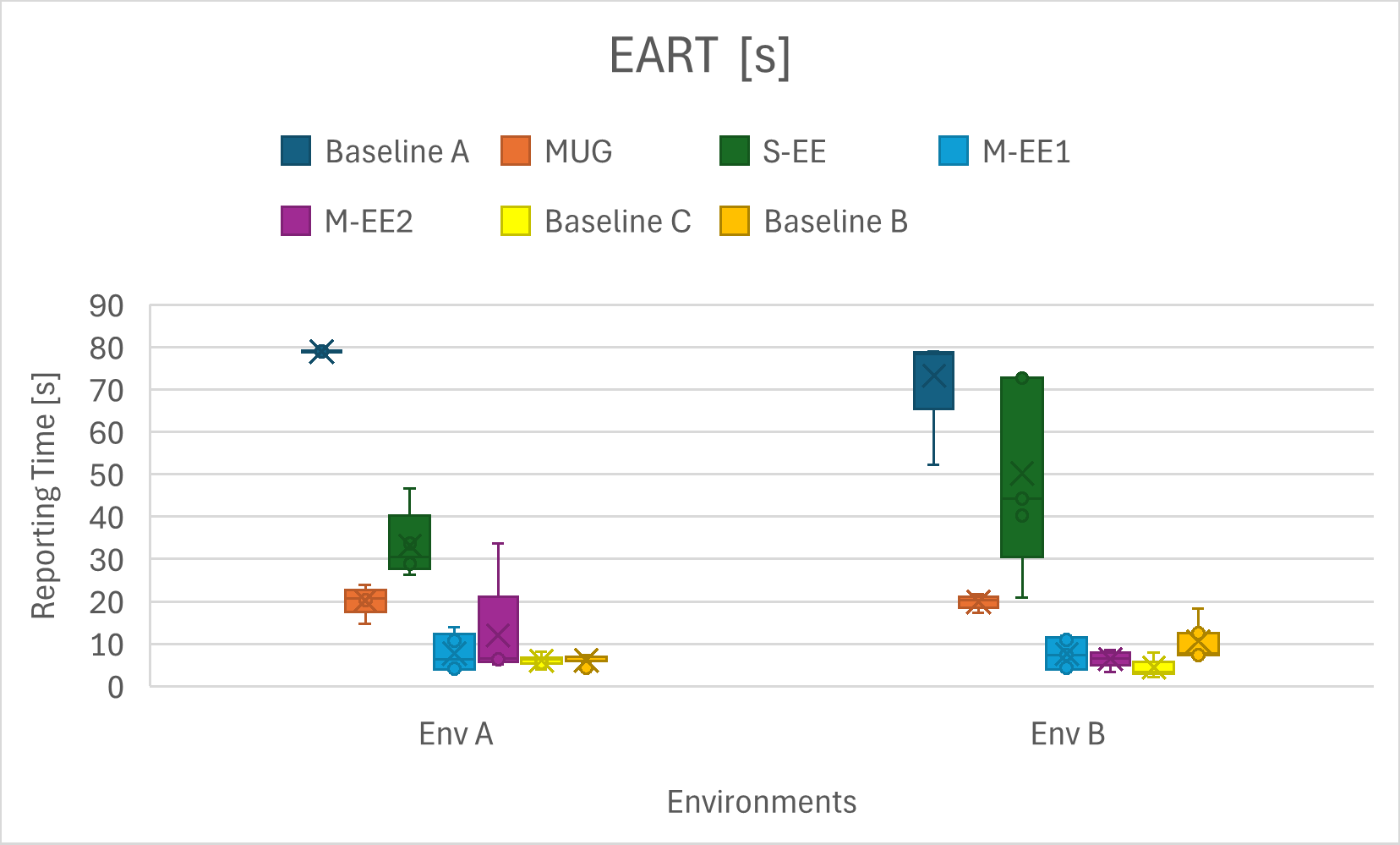}
    \caption{Results for metric $EART$.}
    \label{EART_metric}
\end{figure}

Fig. \ref{ETR_metric} and Fig. \ref{EART_metric} confirm that all cooperative approaches outperformed the Baseline A regarding agent reporting. The majority of the Baseline A tests did not result in any reporting opportunity. The respective $EART$ results are thus equal to the mission length.

The comparison between cooperation strategies using mobile agents, and particularly M-EE2, demonstrates the impact of prior environment knowledge on reducing the variability in the results of this metric. The MUG use case verified consistent reporting in both environments, albeit at a lower level than the M-EE use cases and Baseline B.

At last, it is important to highlight the correlation between the distinct characteristics of M-EE2 and Baseline C, and the different POC distributions in Env A and Env B. M-EE2 had the best performance on the $ETR$ metric, followed by Baseline C, with a performance gap of 24\%. Env B verified the reverse conclusions, as the NART motion restriction under a smaller area of high interest enabled the combination of global reports with further opportunistic pairwise interactions. M-EE2 underperformed Baseline C by 46.97\% in the metric $ETR$. Baseline C achieved the best performance for $EART$ in both environments, followed by M-EE2, with a performance gap of 49.09\% in Env B and 2.3\% in Env A. The reports in Baseline C were more condensed near the time of the global rendezvous. The pairwise reports for the M-EE and MUG use cases are more dispersed throughout the mission due to the individual rewards, which account for the individual $VoM$ of the agents.

\subsection{Global Situational Awareness}\label{goal4}

The ability of the NART to accumulate a global situational awareness is analysed through the $ETAK$ and $EIK$ metrics.

\begin{figure}[htp]
    \centering
    \includegraphics[scale = 0.6]{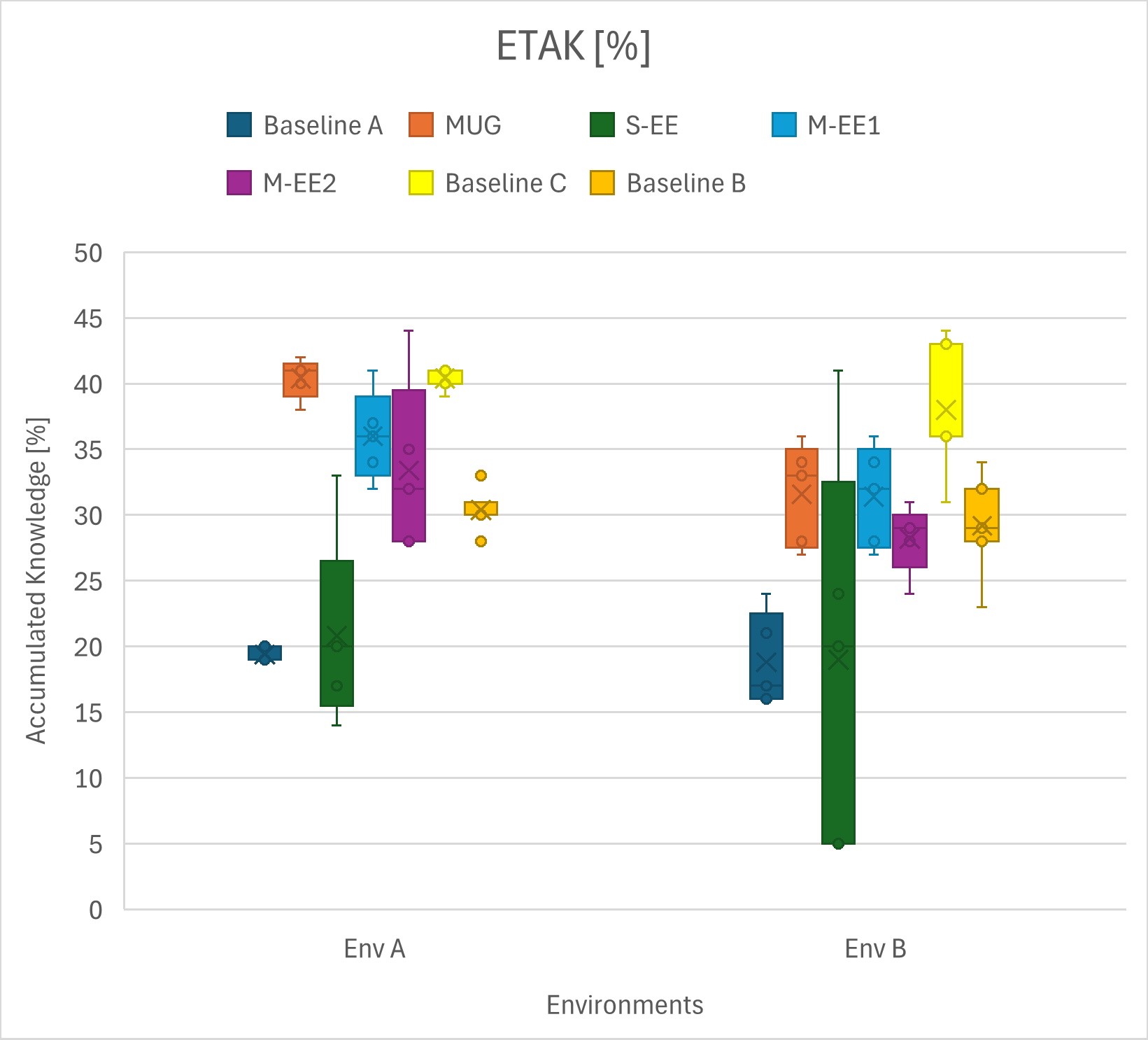}
    \caption{Results for metric $ETAK$.}
    \label{ETAK_metric}
\end{figure}

The metric $ETAK$ presented in Fig. \ref{ETAK_metric} quantifies the global situational awareness of the NART agents. Each explorer agent can improve its situational awareness by visiting and revisiting cells along its own path, or through cooperation with other agents. On the other hand, the external entities can only improve their global situational awareness through cooperation.

The $ETAK$ value of a NART reflects the ability to balance reporting and exploration tasks, as both contribute to this metric. As the individual $EAK$ tends to increase with time, a data exchange at the end of the mission might enable a larger $EAK$ increase than a data exchange in the first steps of the mission, particularly after a long time interval without meetings. Nevertheless, sparse reporting may provide a consistent $EAK$ increase and improved global situational awareness during the mission.

Across different use cases, the UAVs individually achieved up to 19\% of $EAK$. These values are obtained, for example, in most Baseline A tests. However, through consistent cooperation, this value increased to 50\% in the M-EE1 use case, 46\% in the MUG use case, 41\% in Baseline C, and 40\% in the M-EE2 use case (in Env A).

Beyond the frequency and time of reporting, $ETAK$ can also reflect the impact of significant discrepancies in individual agent reports. Agents can have a lower $EAK$ value due to restricted motion or inefficient exploration, which in turn reduces the metric results. For example, while static EE, under adequate EE positioning, can provide an increased number of overall reports between NART agents. It, however, does not ensure that the reports are evenly distributed across the agents, resulting in a low $ETAK$ value. On the other hand, although the MUG use case estimates a smaller number of reports, it provides a more even distribution of accumulated knowledge among all agents and achieves top performance in this metric.

\begin{figure}[!t]
    \centering
    \includegraphics[scale = 0.6]{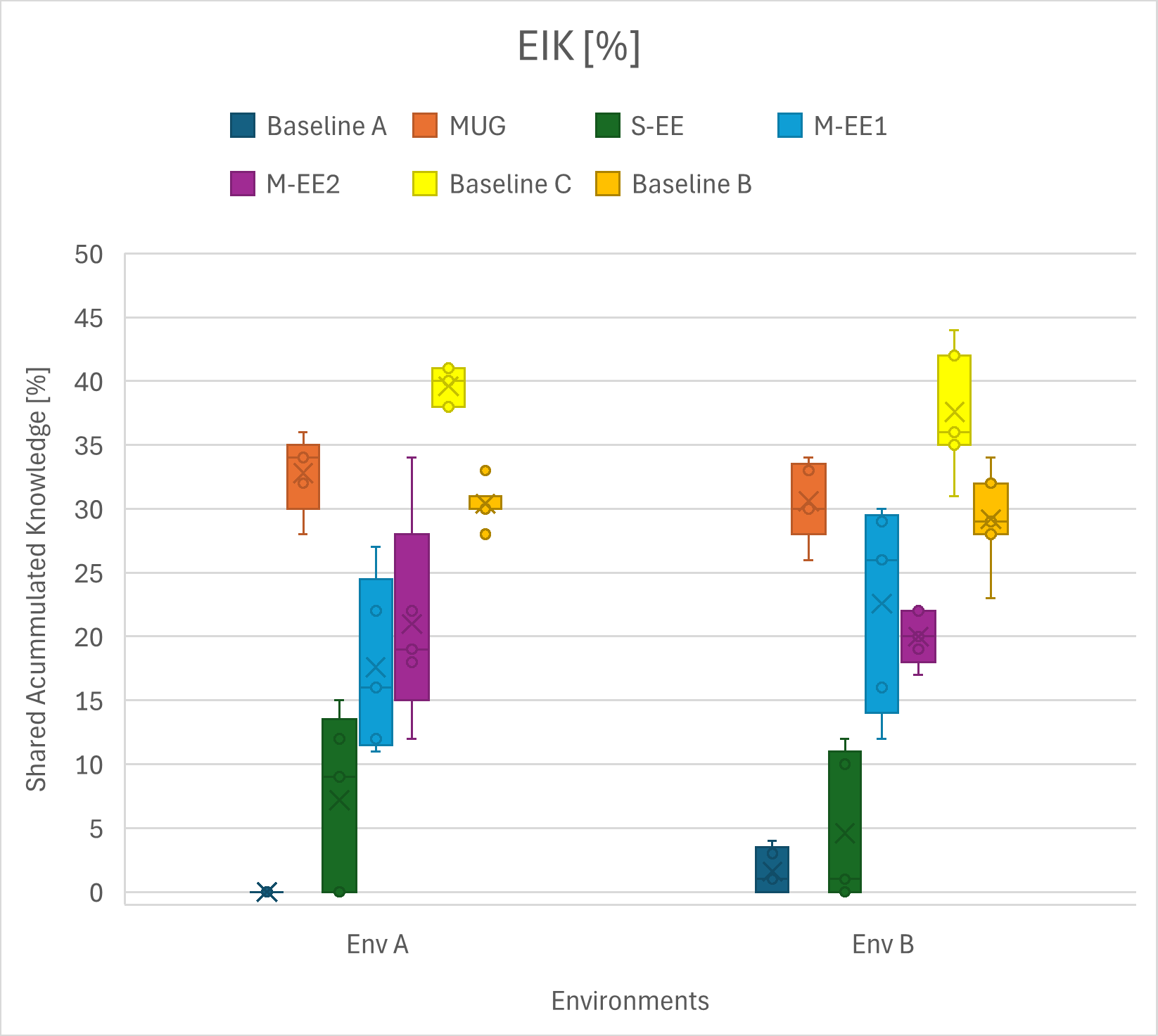}
    \caption{Results for metric $EIK$.}
    \label{SGSA_metric}
\end{figure}

The metric $EIK$ presented in Fig. \ref{SGSA_metric} quantifies the expected situational awareness intersection of the NART agents.
The agents are not restricted to cooperate with all their neighbours.  Thus, a high performance at the NART reporting task does not necessarily correspond to a high $EIK$ value. If agents visit the same cells at different times, the NART can have a non-null $EIK$ value even without reports. Achieving a high level of knowledge intersection between individual agent situational awareness requires all agents to communicate with each other, or for the information regarding the environment to be relayed to all NART agents.

The MUG use case and Baseline C were the best-performing strategies in this metric for both environments. Both strategies have very similar $ETAK$ results, with a performance gap of less than 0.1\% in Env A and 2.7\% in Env B, with Baseline C achieving the best performance. However, the combination of global reports and opportunistic pairwise interactions in Baseline C fostered the sharing of individual $EAK$ by all NART agents, thereby yielding an equivalent $EIK$ metric that outperformed the MUG use case by 15\% in Env A and 16.67\% in Env B.

Despite high $EAK$ results across most M-EE strategies, this performance did not translate into $EIK$ improvements. The support of a mobile external entity, introduced similarly as a data mule, helped to increase the $EAK$ of some agents in the NART. As the information has not been relayed to all agents, this has resulted in greater disparity in global situational awareness and, consequently, lower performance on the $EIK$ metric.

Baseline B underperformed the MUG use case in both environments, with a performance loss of 25.9\% for $ETAK$ and 10.6\% for $EIK$ in Env A, and 24\% and 11.3\% in Env B, respectively. M-EE1 and M-EE2 use cases also outperformed Baseline B in Env B. However, the difference between the results of the metrics $ETAK$ and $EIK$ is below 0.1\% on average in both environments. It is important to note that the results for $ETAK$ and $EIK$ under Baseline B, as in Baseline C, indicate greater ability to transmit information gathered by individual agents to the remaining NART agents. All other strategies show a larger discrepancy between the averaged $ETAK$ and $EIK$ results. In both environments, the second-ranked strategy is the MUG use case with 7\% in Env A and 5\% in Env B.

It is also possible to draw further conclusions regarding the ability of agents to contribute to a global situational awareness under intermittent connectivity. Considering the percentage that $EIK$ represents in the $ETAK$ results, on average, the intersected knowledge in the MUG use case represents 85.7\% of their total accumulated knowledge in Env A and 82.9\% in Env B. Regarding the use cases with a mobile EE, the addition of a role-based behaviour loop improved the ability of agents to share their accumulated knowledge when a non-uniform POC distribution was considered. While the M-EE2 had the second-best performance (81.2\%) in Env B, M-EE1 had the second-best performance (59.4\%) in Env A. A similar analysis to the S-EE use case results shows that in Env A, $EIK$ represents 45\% of their $ETAK$ value. At last, in the Baseline A approach, $EIK$ accounts for 4.1\% of the NART $ETAK$ in Env B and 0\% for Env A.

\subsection{Impact of $w_1$ and $w_2$ in NART Performance}

The impact of $w_1$ and $w_2$ in the cooperative incentives and resulting group performance, will vary for different NART topologies. For this analysis, a M-EE1 use case was considered in Env A. Each combination of weights was tested three times. The average results and respective standard deviations are presented in Table \ref{tab_w}.

By tuning the weights $w_1$ and $w_2$, the NART can achieve better performance across distinct mission goals. Prioritising agent exploration leads to the higher exploration performance, namely in metrics $E$, $TPOC$ and $EP$. The pairing $w_1 = 0.7$ and $w_2= 0.3$ achieved the best results in metrics $E$ and $TPOC$. Metric $EP$ achieved the best result with the $w_1 = 0.3$ and $w_2= 0.7$ combination. In the performed tests, the exploration was mostly jeopardised only when $w_1=0$. The mobile external entity has a fixed trajectory near the perimeter of the area to explore. Promoting cooperation with this entity also indirectly encourages exploration. However, exclusively reporting to the external entity ($w_2=1$) can lead to overly constrained agent trajectories.

\begin{table*}[!htbp]
\centering
\caption{Impact of $w_1$ and $w_2$ on M-EE1 Performance in Env A.}
\label{tab_w}
\begin{adjustbox}{max width=0.6\textwidth}
\begin{tabular}{|c|c|c|c|c|c|c|c|c|}
\hline
\multicolumn{2}{|c|}{\textbf{Test}} & \multicolumn{7}{c|}{\textbf{Metrics}} \\
\hline
$w_1$ & $w_2$ & $E$ & $TPOC$ & $EP$ & $ETR$ & $EART$ & $ETAK$ & $EIK$ \\
\hline
1 & 0 & \makecell{0.2338 \\ ($\pm$ 0.0028)} & \makecell{32.77 \\ ($\pm$ 0.53)} & \makecell{75.7 \\ ($\pm$ 0.58)} & \makecell{26.33 \\ ($\pm$ 13.79)} & \makecell{12.71 \\ ($\pm$ 3.04)} & \makecell{32.3 \\ ($\pm$ 0.6)} & \makecell{6.7 \\ ($\pm$ 11.5)} \\
\hline
0.7 & 0.3 & \makecell{0.2555 \\ ($\pm$ 0.0227)} & \makecell{34.82 \\ ($\pm$ 0.79)} & \makecell{68.3 \\ ($\pm$ 21.22)} & \makecell{31.33 \\ ($\pm$ 9.08)} & \makecell{8.19 \\ ($\pm$ 2.54)} & \makecell{33.7 \\ ($\pm$ 11.6)} & \makecell{15.3 \\ ($\pm$ 7.6)} \\
\hline
0.5 & 0.5 & \makecell{0.2367 \\ ($\pm$ 0.0064)} & \makecell{33.63 \\ ($\pm$ 1.18)} & \makecell{77.0 \\ ($\pm$ 3.46)} & \makecell{28.67 \\ ($\pm$ 2.08)} & \makecell{9.16 \\ ($\pm$ 1.01)} & \makecell{37.0 \\ ($\pm$ 1.7)} & \makecell{12.0 \\ ($\pm$ 11.3)} \\
\hline
0.3 & 0.7 & \makecell{0.2423 \\ ($\pm$ 0.00284)} & \makecell{34.42 \\ ($\pm$ 0.38)} & \makecell{80.3 \\ ($\pm$ 1.54)} & \makecell{71.3 \\ ($\pm$ 6.43)} & \makecell{4.42 \\ ($\pm$ 1.21)} & \makecell{43.3 \\ ($\pm$ 2.5)} & \makecell{31.7 \\ ($\pm$ 2.9)} \\
\hline
0 & 1 & \makecell{0.2187 \\ ($\pm$ 0.0202)} & \makecell{30.29 \\ ($\pm$ 3.56)} & \makecell{68.0 \\ ($\pm$ 8.89)} & \makecell{45.67 \\ ($\pm$ 18.18)} & \makecell{10.48 \\ ($\pm$ 1.51)} & \makecell{32.0 \\ ($\pm$ 5.6)} & \makecell{13.7 \\ ($\pm$ 7.5)} \\
\hline
\end{tabular}
\end{adjustbox}
\end{table*}

On the other hand, a higher weight $w_2$ results in a higher number of reports, $ETR$. The $ETR$ and the user-defined battery life of an agent influence its average reporting time. More demanding trajectories (i.e., with a higher energy cost) can be shorter than those with lower energy demand to comply with the limited agent energy. In this case, a similar number of reports can yield different $EART$ results.

An external entity can act as a data mule, improving data sharing among agents. Exploration also increases the $ETAK$ results. Thus, the balance between cooperating with the M-EE and exploring different areas of the environment is key to improving NART global situational awareness. This impact was found particularly when $w_1 = 0.3$ and $w_2= 0.7$ and  when $w_1 = 0.7$ and $w_2= 0.3$. 


\subsection{Ablation Study}

An ablation study was also conducted to clarify the effects of the different components of the proposed cooperative strategy on NART performance, namely, the communication-shaped rewards and the proposed communication model. The results presented in Table \ref{ablationt} consider 3 trials performed in the same conditions as stated in Section 4. 

\begin{table*}[!htbp]
\centering
\caption{Ablation Study, considering a NART composed of 3 UAVs in Env A.}
\label{ablationt}
\footnotesize
\begin{adjustbox}{max width=0.8\textwidth}
\begin{tabular}{|c|c|c|c|c|c|c|c|c|c|c|}
\hline
\multicolumn{4}{|c|}{\textbf{Tests}} & \multicolumn{7}{c|}{\textbf{Metrics}} \\
\hline
Number & Exploration & Reward & Proposed CM & $E$ & $TPOC$ & $EP$ & $ETR$ & $EART$ & $ETAK$ & $EIK$ \\
\hline
1 &Yes & No & Yes & \makecell{0.2631 \\($\pm$ 0.0005)}  & \makecell{37.84 \\($\pm$ 0.10)} & \makecell{0.927 \\($\pm$ 0.006)} & \makecell{0 \\($\pm$ 0)} &  \makecell{79 \\($\pm$ 0)} &  \makecell{0.197 \\($\pm$ 0.006)} &  \makecell{0 \\($\pm$ 0)} \\
\hline
2 &Yes & $CSI$ & Yes &  \makecell{0.1731 \\($\pm$ 0.018)} &  \makecell{23.65 \\($\pm$ 3.10)} &  \makecell{0.453 \\($\pm$ 0.085)} &  \makecell{418 \\($\pm$ 24.33)} &  \makecell{0.520 \\($\pm$ 0.021)} &  \makecell{ 0.367 \\($\pm$ 0.045)} &  \makecell{0.367 \\($\pm$ 0.045)} \\
\hline
3 &Yes & $VoM$ & Yes &  \makecell{0.1889 \\($\pm$ 0.0014)} &  \makecell{28.12 \\($\pm$ 0.16)} &  \makecell{0.673 \\($\pm$ 0.006)} &  \makecell{1.33 \\($\pm$ 1.15)} &  \makecell{72 \\($\pm$ 0)} &  \makecell{0.21 \\($\pm$ 0.06)} &  \makecell{0 \\($\pm$ 0)} \\
\hline
4 &Yes & $CSI$ and $VoM$ & No &  \makecell{0.2164 \\($\pm$ 0.006)} &  \makecell{29.89 \\($\pm$ 0.84)} &  \makecell{0.660 \\($\pm$ 0.0.026)} &  \makecell{6.67 \\($\pm$ 3.06)} &  \makecell{42.85 \\($\pm$ 21.50)} &  \makecell{0.380 \\($\pm$ 0.044)} &  \makecell{0.303 \\($\pm$ 0.06)}\\
\hline
5 &Yes & $CSI$ and $VoM$ & Yes &  \makecell{0.2410 \\($\pm$ 0.0056)}&  \makecell{33.66 \\($\pm$ 0.95)} &  \makecell{0.800 \\($\pm$ 0.035)} &  \makecell{11.33 \\($\pm$ 1.15)} &  \makecell{21.96 \\($\pm$ 1.75)} &  \makecell{0.403 \\($\pm$ 0.021)} &  \makecell{0.313 \\($\pm$ 0.031)}\\
\hline
\end{tabular}
\end{adjustbox}
\vspace{1mm}\\
\noindent\footnotesize\textit{Note:} Communication Model (CM).
\end{table*}

Test 1 corresponds to Baseline A, as it considers only exploration, and Test 5 corresponds to the proposed strategy (MUG use case). Tests 2 to 4 present variations of the proposed strategy.
The combination of the proposed communication-shaped rewards and the communication model verified a compromise between frequent reporting and increased NART global situational awareness, while minimising their impact on exploration. The highest exploration performance was achieved in Test 1. Test 2 had the highest reporting performance. Test 5 had the overall highest global situational awareness.

When integrated simultaneously, the proposed reward components play different roles in shaping agent rewards and overall NART capabilities. While the $VoM$ supports flexible time intervals between rendezvous, the $CSI$ reflects the feasibility of those rendezvous within the capabilities of the communication system (e.g., radio technology and physical phenomena). 
This way, it is possible to balance the need and ability of an agent to communicate, thereby promoting more flexible cooperation with other NART agents. The proposed strategy helps reduce the impact on exploration while also achieving higher reporting and global situational awareness. 

\subsubsection*{Exploration}
In the environment tested, high communication indexes correspond to the spatial approximation of all NART agents. This spatial proximity, required for high behavioural rewards in Test 2, also leads to a significant NART motion restriction, resulting in lower exploration performance. In Test 2, metrics $E$, $TPOC$, and $EP$ showed a performance reduction of 28.18\%, 29.76\% and 43.33\%compared to Test 5.

In Test 3, rewards solely determined by $VoM$ discourage reporting, as the maximum reward is achieved when $VoM$ is at its maximum. This lack of reporting also includes avoiding opportunistic rendezvous, which at most occur in the first steps of the mission. The resulting dispersed trajectories yield a slight exploration performance of 8.387\% in metric $E$, 15.91\% in metric $TPOC$, and 32.67\% in metric $EP$, compared to Test 2. However, its performance is still lower than Test 5 by 21.62\%, 16.47\% and 15.83\% as well as Test 1 by 34.22\%, 37.51\% and 51.08\% in metrics $E$, $TPOC$ and $EP$, respectively.

\subsubsection*{Reporting}
Considering only $CSI$ as a behavioural reward in Test 2 results in higher NART connectivity. The rewards increase as all agents verify a large individual $CSI$. This behaviour results in high reporting ability (i.e., high $ETR$ and low $EART$), the highest among all combinations tested. Test 3 verifies the opposite, with minimal reporting.

The performance of Test 2 on the $ETR$ metric largely surpassed that of Test 5 by 3588.24\%. The $EART$ metric was 97.65\% lower, representing shorter time intervals between rendezvous. Test 3 underperformed Test 5 with lower $ETR$ results (88.23\%) and higher $EART$ (227.92\%) results.

\subsubsection*{Global Situational Awareness}
In Test 2, the frequent reports support the dissemination of the estimated knowledge gained by each agent to all agents and, therefore, the global situational awareness.  $ETAK$ results in Test 2 was 9.09\% lower than in Test 5. However, as the $ETAK$ and $EIK$ results are the same in Test 2, the reporting increases $EIK$ performance by 17.02\% compared to Test 5, where lower reports were estimated.

With minimal reporting, Test 3 shows a large impact on global situational awareness performance, even though it shows higher exploration performance than Test 2. The $EIK$ metric yielded 0, indicating that considering only the VoM as a reward penalises the transmission of the expected individual accumulated knowledge to all NART agents. The $ETAK$ metric surpassed the result in Test 1 by 5.08\% but underperformed the result in Test 5 by 48.76\%.

\subsubsection*{Impact of the Communication Model in NART Performance}

Comparing the results for Tests 4 and 5, it is possible to conclude that abstracting the capabilities and limitations of radio technology can jeopardise mission planning and NART resilience. Regarding exploration performance, Test 4 performs worse across all metrics compared to Test 5: 20.23\% in the $E$ metric, 11.19\% in the $TPOC$ metric, and 17.5\% in the $EP$ metric. Reporting ability in Test 4 was also lower than in Test 5, with a lower estimated number of reports (41.18\%) and longer reporting times (95.16\%). Global Situational Awareness was also reduced, as Test 4 obtained $ETAK$ and $EIK$ results 5.79\% and 3.19\% lower than in Test 5.

\section{Communication and Motion Coordination Awareness in the Proposed Approach}\label{CMCA_App}

In \cite{CMAMCA}, a taxonomy for NART awareness was proposed. Works are categorised based on the level of detail they assign to the communication and motion coordination components of NARTs. The awareness level of communication is identified by CMA while the awareness level of motion coordination is identified by MCA. Communication strategy and connectivity requirements can be abstracted (CMA-A), set prior to mission start (CMA-B) or adapted during the mission development (CMA-C). On the other hand, motion coordination of NART agents can be disregarded (i.e., considering only independent behaviours, MCA-A), considered under predefined strategies (MCA-B) or adaptive approaches (MCA-C). Sublevels (e.g., MCA-B1 and MCA-B2) further distinguish different the component awareness.

The motion of the agents is optimised prior to mission start. The proposed strategy balances two goals: staying within communication range to exchange data and spreading out to explore the environment. The motion coordination awareness implemented with this work can be considered as MCA-B2. 

This work does not consider permanent connectivity and cooperation between the agents is not guaranteed to occur. The communication restrictions are given according to wireless channel modelling and limitations of a user-selected radio technology. It combines the restrictions that provide data exchanges with the need to sparse agents in the environment, resulting in a NART with a highly flexible topology. The current work considered the evaluation of the NART performance at the end of the optimisation, based on estimated communication requirements. The Communication Awareness in this work corresponds to a CMA-B2 level.

\subsubsection*{Trade-off between Communication and Motion Coordination}

\begin{figure}[!t]
    \centering
    \begin{subfigure}[b]{\columnwidth}
        \centering
        \includegraphics[width=\linewidth]{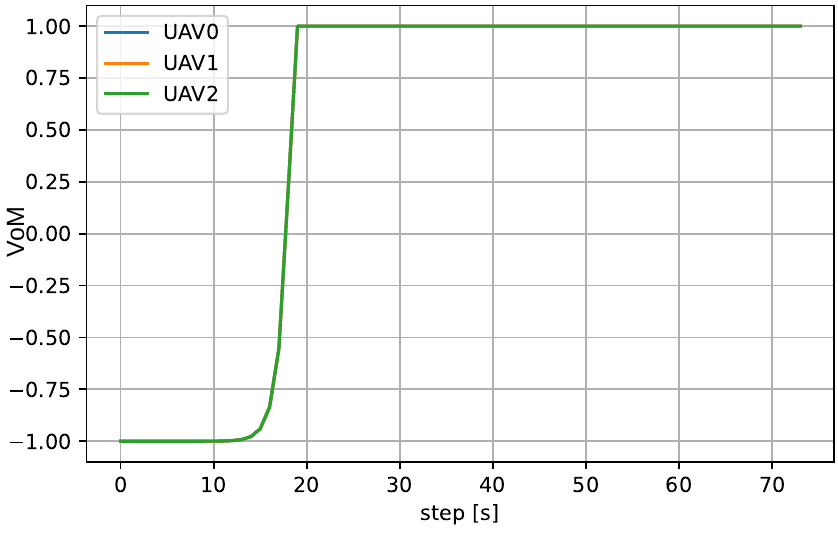} 
        \caption{ }
        \label{VOMSEE2}
    \end{subfigure} 
    \begin{subfigure}[b]{\columnwidth}
        \centering
        \includegraphics[width=\linewidth]{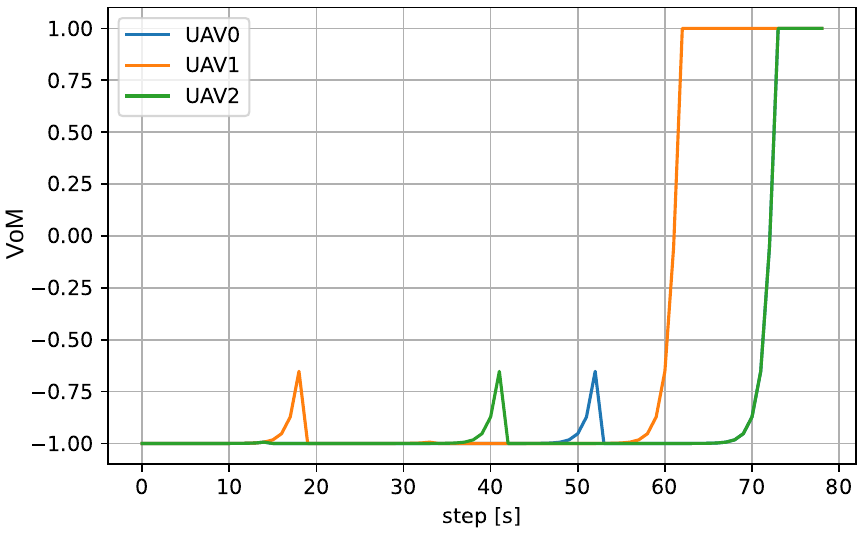} 
        \caption{ }
        \label{VOMMUG4}
    \end{subfigure}
    \caption{$VoM$ results for a test without reports (\ref{VOMSEE2}) and with cooperation and frequent reports (\ref{VOMMUG4}).}
    \label{results21}
\end{figure}

The chosen use cases combine different compositions and restrictions of NARTs. The trade-offs found through the joint consideration of exploration and reporting become particularly visible through the analysis of the VoM of the agents, as both time and agent reports influence the VoM evolution and the agent behaviour loops. Due to minimal, if any, reporting, the VoM is maintained at its maximum value for the majority of the mission in a non-cooperative mission, as illustrated in Fig. \ref{VOMSEE2}, or mostly at minimum values in a mission with persistent reporting, as presented in Fig. \ref{VOMMUG4}.

\begin{figure}[!t]
    \centering
    \begin{subfigure}[b]{\columnwidth}
        \centering
        \includegraphics[width=\linewidth]{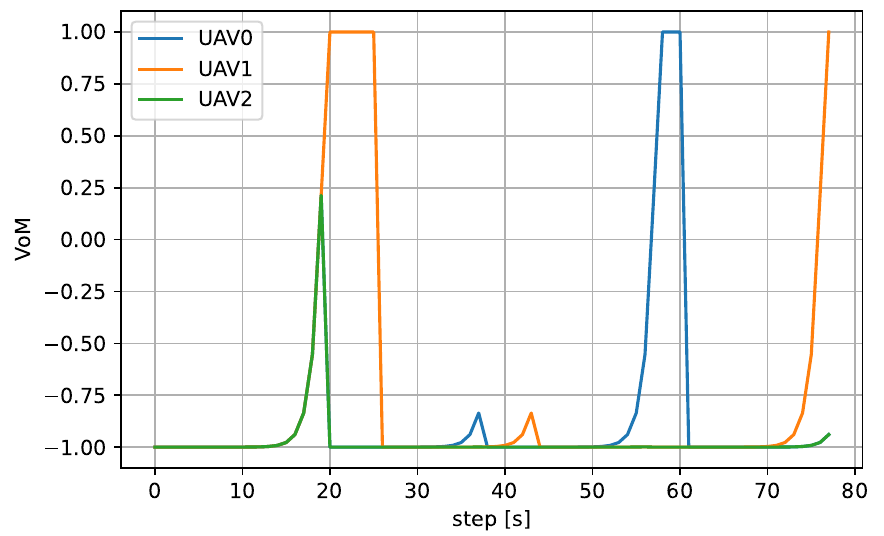} 
        \caption{ }
        \label{results2_a}
    \end{subfigure} 
    \begin{subfigure}[b]{\columnwidth}
        \centering
        \includegraphics[width=\linewidth]{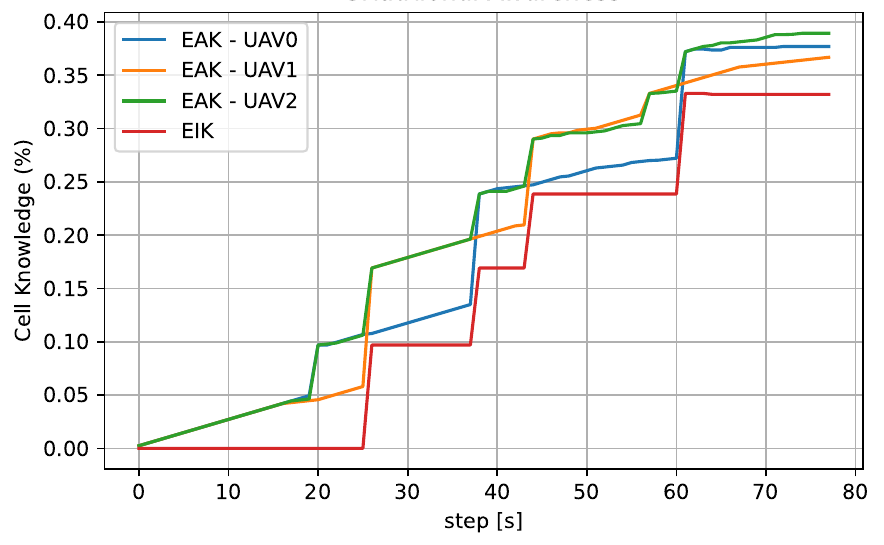} 
        \caption{ }
        \label{results2_b}
    \end{subfigure}
    \caption{Dynamic trade-off: $VoM$ \ref{results2_a} and $EAK$ \ref{results2_b} metrics (MUG Use Case).}
    \label{results2}
\end{figure}

On the other hand, missions with frequent yet sparse reports made the trade-off more challenging. As the proposed approach does not impose hard constraints on UAV roles, emerging UAV behaviours include seamlessly exchanging between exploring (high VoM) and reporting (low VoM) tasks in a desynchronised pattern. Fig. \ref{results2_a} illustrates one of the patterns that balances individual situational awareness, reporting, and data relaying. The consequent gradual increase in the NART situational awareness throughout the mission is presented in Fig. \ref{results2_b}. This contrasts with the evolution of the situational awareness metrics of the Baseline A approach. The above mentioned trade-off was found in both mission environments. The MUG use case also leveraged prior information about the environment and the consequent definition of high-interest areas to improve the $TPOC$ metrics and reduce the gap to the Baseline A approach.

\section{Conclusion}\label{conclusion}

This work proposes an offline trajectory optimisation with rendezvous incentives under several NART limitations, including sensor uncertainty, limited energy, intermittent communication and prior environment information. Cooperation is considered a communication-dependent event, subject to the limitations of a user-selected radio technology. The agent behaviour follows a high-level behaviour loop, with soft restrictions regarding agent roles and dynamic rewards for cooperation. The tool presented provides optimised paths for the NART agents, based on their limitations and mission goals. It considers the impact of data sharing between NART agents to increase their global situational awareness and provides additional reporting opportunities. Upon NART deployment, these opportunities can be utilised for transmitting additional mission and environment updates, as well as for mission replanning if required, thereby enhancing NART resilience in dynamic and partially unknown environments. Distinct NART constitutions were tested, including a multi-UAV group working independently and with the support of static and mobile external entities. 

A ground sensing coverage use case was chosen to illustrate the capabilities of the proposed tool. The obtained results highlighted specific strengths of each approach. The choice of the best-suited NART strategy for a particular mission will depend on the available resources (mobile and static agents) and mission requirements (prioritising exploration, reporting, or a balance between them).  

The non-cooperative approach chosen demonstrated the best performance in maximising the probability of finding targets in the environment. The second-best performance in this task was achieved by the independent multi-UAV group, with a performance gap under 10\%. The support of external entities fostered reporting between agents and increased situational awareness. The most reliable reports were achieved with the support of a mobile external entity, reducing reporting latency by over 90\% compared to the Baseline A, at the cost of a reduced performance in the exploration task (up to 13\% in the $E$ metric). While the situational awareness in the Baseline A increased up to 19\%, cooperative approaches achieved 
results up to 40\% (M-EE2), 46\% (MUG) and 50\% (M-EE1). The independent multi-UAV group reached a balanced and consistent performance across all metrics. It was also highlighted for its ability to create a high intersected situational awareness among NART agents.

Two cooperative Baselines were considered. Both combined global rendezvous with fixed time intervals between them. One abstracted the characteristics of radio technologies (Baseline B) and the second included the proposed communication model (Baseline C). These strategies led to greater motion restriction and, consequently, lower exploration performance. It also promotes data exchange among all agents, with high reporting performance. For Baseline C, it also boosted global situational awareness performance. Abstracting agent connectivity conditions can also overly restrict the motion of the agents upon reporting and unnecessarily jeopardise mission performance.

The proposed approach with pairwise interaction and dynamic cooperative incentives can provide a more versatile approach to address communication limitations in NARTs. As the communication model adapts the expected communication range to specific user-defined radio technologies, manual adjustment of the distance between agents to accommodate different radio technology limitations is not required. The modular structure of the proposed model also facilitates the integration of additional mechanisms to enhance the communication awareness of the NART. The straightforward output of the expected link connectivity enhances the interpretability of NART communication and facilitates its integration into various NART applications.

To further understand the potential of communication-aware trajectory optimisation within different NART compositions, future work will address path planning with explicit rendezvous optimisation. The extension of the current optimisation to a larger set of currently user-defined parameters can also provide additional insights. Future work will also focus on how the degree of shared situation awareness will impact the performance of online multi-agent decision systems (e.g., based on Multi-Agent Reinforcement Learning, MARL).

\section*{Acknowledgements}
This work was supported by the Portuguese Foundation for Science and Technology (FCT) under Grant 2023.04842.BD and by LARSyS FCT funding (DOI: 10.54499/LA/P/0083/2020, 10.54499/UIDP/50009/2020, and 10.54499/UIDB/50009/2020). This work is also funded by national funds through FCT — Fundação para a Ciência e a Tecnologia, I.P., under projects/supports :

\begin{itemize}
    \item UID/6486/2025 \\(https://doi.org/10.54499/UID/06486/2025)
    \item UID/PRR/6486/2025 \\(https://doi.org/10.54499/UID/PRR/06486/2025)
    \item UID/PRR2/06486/2025 \\(https://doi.org/10.54499/UID/PRR2/06486/2025)
\end{itemize}

\bibliographystyle{elsarticle-num}
\bibliography{bib}

\end{document}